\documentclass[journal]{IEEEtran} 

\usepackage{graphics} %
\usepackage{epsfig} %
\usepackage{graphicx}
\usepackage{xcolor} %
\usepackage{capt-of}

\usepackage{amsmath} %
\usepackage{amssymb}  %
\usepackage{amsfonts}
\usepackage{mathtools}%
\usepackage{units}
\usepackage{algorithmic}
\usepackage{algorithm}
\usepackage{array}
\usepackage{xfrac}

\usepackage{booktabs}
\usepackage{multirow}
\usepackage{tabularx}
\usepackage{multirow}

\usepackage{hyperref}
\usepackage{balance}
\usepackage{cite}
\usepackage{textcomp}
\usepackage{stfloats}
\usepackage{url}
\usepackage{verbatim}

\usepackage{tikz}
\usepackage{pgfplots}

\usepackage{pgfplotstable}
\usepackage{circuitikz}
\usetikzlibrary{bending, plotmarks, shapes, pgfplots.statistics, pgfplots.colorbrewer, positioning, arrows.meta, matrix}

\pgfplotsset{scaled y ticks=false}
\pgfplotsset{compat=1.16,
    width = 7cm,
    height = 3.5cm,
    title style = {yshift=-6pt},
    xlabel shift = -3pt,
    ylabel shift = -0pt,
    cycle list={{matlab1},{matlab2},{matlab3},{matlab9},{matlab4},{matlab5},{matlab6},{matlab7},{matlab8}},
    legend columns = 1,
    legend cell align={left},
    legend style ={
        draw = gray,
        fill opacity=0.8,
        text opacity=1.0,
        draw opacity=1.0,
    },
    xmajorgrids,
    ymajorgrids,
    scale only axis,
} 

\newcommand{\cfr}[0]{\ensuremath{\mathcal{C}}} %

\newcommand{\rotateRPY}[3]%
{   \pgfmathsetmacro{\rollangle}{#1}
    \pgfmathsetmacro{\pitchangle}{#2}
    \pgfmathsetmacro{\yawangle}{#3}

    \pgfmathsetmacro{\newxx}{cos(\yawangle)*cos(\pitchangle)}
    \pgfmathsetmacro{\newxy}{sin(\yawangle)*cos(\pitchangle)}
    \pgfmathsetmacro{\newxz}{-sin(\pitchangle)}
    \path (\newxx,\newxy,\newxz);
    \pgfgetlastxy{\nxx}{\nxy};

    \pgfmathsetmacro{\newyx}{cos(\yawangle)*sin(\pitchangle)*sin(\rollangle)-sin(\yawangle)*cos(\rollangle)}
    \pgfmathsetmacro{\newyy}{sin(\yawangle)*sin(\pitchangle)*sin(\rollangle)+ cos(\yawangle)*cos(\rollangle)}
    \pgfmathsetmacro{\newyz}{cos(\pitchangle)*sin(\rollangle)}
    \path (\newyx,\newyy,\newyz);
    \pgfgetlastxy{\nyx}{\nyy};

    \pgfmathsetmacro{\newzx}{cos(\yawangle)*sin(\pitchangle)*cos(\rollangle)+ sin(\yawangle)*sin(\rollangle)}
    \pgfmathsetmacro{\newzy}{sin(\yawangle)*sin(\pitchangle)*cos(\rollangle)-cos(\yawangle)*sin(\rollangle)}
    \pgfmathsetmacro{\newzz}{cos(\pitchangle)*cos(\rollangle)}
    \path (\newzx,\newzy,\newzz);
    \pgfgetlastxy{\nzx}{\nzy};
}

\tikzset{
  drone/.pic={
    \begin{scope}[scale=#1, yshift=4cm]
      \filldraw (-3.9, -3.5) rectangle (-3.3, 0);
      \filldraw (3.9, -3.5) rectangle (3.3, 0);
      \filldraw (-4.1, 0) -- (-4.2, 0.5) -- (4.2, 0.5) -- (4.1, 0) -- cycle;
      \filldraw (-4.6, 0.5) rectangle ++ (2, 0.8);
      \filldraw (4.6, 0.5) rectangle ++ (-2, 0.8);
      \filldraw (-3.7, 1.3)  rectangle ++ (0.2, 0.5);
      \filldraw (3.7, 1.3)  rectangle ++ (-0.2, 0.5);
      \filldraw (-1.5, -3.0) rectangle (1.5, 0);
      \begin{scope}[xshift = -3.6cm, yshift = 1.8cm, xscale = 3.2, yscale = 0.5]
        \draw[fill, domain=0:360, samples=200, smooth, variable=\t]  
          plot ({cos(\t)}, {sin(2*\t)/2});
      \end{scope}
      \begin{scope}[xshift = 3.6cm, yshift = 1.8cm, xscale = 3.2, yscale = 0.5]
        \draw[fill, domain=0:360, samples=200, smooth, variable=\t]  
          plot ({cos(\t)}, {sin(2*\t)/2});
      \end{scope}
    \end{scope}
  }
}

\DeclareMathOperator*{\argmin}{arg\,min}

\newcolumntype{C}{>{\centering\arraybackslash}X}
\newcolumntype{x}[1]{>{\centering\let\newline\\\arraybackslash\hspace{0pt}}p{#1}}
\definecolor{matlab1}{rgb}{0.00000,0.44700,0.74100}
\definecolor{matlab2}{rgb}{0.85000,0.32500,0.09800}
\definecolor{matlab3}{rgb}{0.92900,0.69400,0.12500}
\definecolor{matlab4}{rgb}{0.49400,0.18400,0.55600}
\definecolor{matlab5}{rgb}{0.4660, 0.6740, 0.1880}
\definecolor{matlab6}{rgb}{0.3010, 0.7450, 0.9330}
\definecolor{matlab7}{rgb}{0.6350, 0.0780, 0.1840}
\definecolor{matlab8}{rgb}{0.8, 0.8, 0}
\definecolor{matlab9}{rgb}{0.6, 0.6, 0.6}
\definecolor{verylightgray}{rgb}{0.98,0.98,0.98}

\pgfdeclarelayer{bg}   
\pgfdeclarelayer{fg}   
\pgfsetlayers{bg,main,fg}

\definecolor{somegray}{rgb}{0.5, 0.5, 0.5}
\newcommand{\darkgrayed}[1]{\textcolor{somegray}{#1}}
\makeatletter
\newcommand*\titleheader[1]{\gdef\@titleheader{#1}}
\AtBeginDocument{%
  \let\st@red@title\@title
  \def\@title{%
    \vskip-2em
    \bgroup\normalfont\large\centering\@titleheader\par\egroup
    \vskip0.5em\st@red@title}
}
\makeatother

\titleheader{\darkgrayed{This paper has been accepted for publication in the IEEE\;Transactions on Robotics (T-RO),\;2026.\;\textcopyright IEEE}}

\newcommand{\rebuttal}[1]{\textcolor{black}{#1}}
\title{\textbf{Low-Latency Event-Based Velocimetry for \\Quadrotor Control in a Narrow Pipe}}

\author{Leonard Bauersfeld and Davide Scaramuzza\\ 
Robotics and Perception Group, University of Zurich, Switzerland\\
\thanks{This work was supported by the European Union’s Horizon Europe Research and Innovation Programme under grant agreement No. 101120732 (AUTOASSESS) and the European Research Council (ERC) under grant agreement No. 864042 (AGILEFLIGHT).}
}

\begin{document}
\maketitle
\thispagestyle{empty}


\begin{abstract}
Autonomous quadrotor flight in confined spaces such as pipes and tunnels presents significant challenges due to unsteady, self-induced aerodynamic disturbances.
Very recent advances have enabled flight in such conditions, but they either rely on constant motion through the pipe to mitigate airflow recirculation effects or suffer from limited stability during hovering.
In this work, we present the first closed-loop control system for quadrotors for hovering in narrow pipes that leverages real-time flow field measurements.
We develop a low-latency, event-based smoke velocimetry method that estimates local airflow at high temporal resolution.
This flow information is used by a disturbance estimator based on a recurrent convolutional neural network, which infers force and torque disturbances in real time.
The estimated disturbances are integrated into a learning-based controller trained via reinforcement learning.
The flow-feedback control proves particularly effective during lateral translation maneuvers in the pipe cross-section. 
There, the real-time disturbance information enables the controller to effectively counteract transient aerodynamic effects, thereby preventing collisions with the pipe wall.
To the best of our knowledge, this work represents the first demonstration of an aerial robot with closed-loop control informed by real-time flow field measurements. 
This opens new directions for research on flight in aerodynamically complex environments. 
In addition, our work also sheds light on the characteristic flow structures that emerge during flight in narrow, circular pipes, providing new insights at the intersection of robotics and fluid dynamics.
\end{abstract}

\vspace*{-4pt}
\section*{Multimedia Material}
\noindent{Video: \url{https://youtu.be/ubHSlYZOeQQ}}

\section{Introduction}
\label{sec:introduction}

In recent years, quadrotors have become increasingly widespread in research~\cite{kaufmann23champion, 2014:DAndrea, 2017:Karydis} and industry~\cite{skydioweb, flyabilityweb}, as their small size and agility make them particularly suitable for many tasks, ranging from inspection and mapping to data collection and search and rescue scenarios.  
In all aforementioned applications, quadrotors operating in narrow environments and reaching locations inaccessible to humans or ground-based robots can potentially be game-changing~\cite{AutoAssess}.  
However, when operating in confined spaces, the induced airflow of the quadrotor causes strong and unsteady aerodynamic disturbances, making safe flight extremely challenging. 
Furthermore, state estimation in such environments is difficult, as they are typically dark and featureless.  
Finally, little is known about the flow configurations that develop in narrow tunnels, since fluid mechanics research has primarily focused on quadrotors flying in free air, both in computational fluid dynamics simulations~\cite{paz2021assesmentcfd, ventura2018high, luo2015novel} and in real-world experiments~\cite{wolf2024volumentricwakeinvestigation, bauersfeld2025roboticsmeetsfluiddynamics}.

\begin{figure}[t!]
    \centering
    \includegraphics{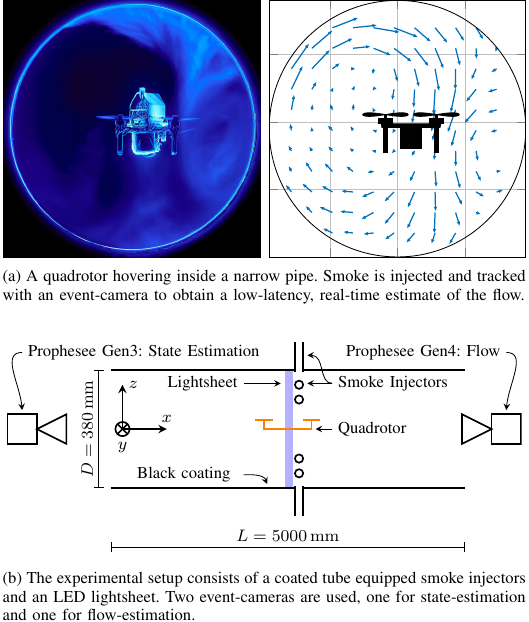}
    \vspace*{-18pt}
    \caption{In this work, we demonstrate how low-latency event-based smoke velocimetry can be used for real-time disturbance estimation to improve the closed-loop control performance of a quadrotor flying inside a narrow pipe.
    }
    \label{fig:1}
\end{figure}

Numerous studies have addressed the challenges associated with flying in confined environments. Initial efforts focused on controlling drones operating within ground effect~\cite{shi2019learnedlanding}, followed by advances enabling agile maneuvers under strong wind conditions~\cite{oconnel2022neuralfly}. Additionally, recent work has shown that aerodynamic disturbances can be exploited to sense nearby obstacles~\cite{ding2023aerodynamiceffect}.  
Only very recently, the first autonomous quadrotor flight with onboard state estimation in narrow pipes was demonstrated, although the vehicle has to keep moving along the pipe in order to partially avoid its ego-airflow~\cite{wang2025autonomousflighttunnel}. In a concurrent work, hovering inside a pipe was demonstrated~\cite{martin2025flyinginducts}, and the authors of the study find that \emph{``the drone has a hard time hovering at a fixed position''}, with positional errors up to \unit[6]{cm}. This brings the drone dangerously close to the pipe walls.

In our work, we go one step further and implement a sparse, low-latency, event-based smoke velocimetry method to measure the airflow in real time, then estimate the aerodynamic disturbances with a recurrent convolutional network and feed this information into an learning-based controller. 
This not only improves the hovering accuracy but also allows more controlled translational motion in the pipe cross-section. 
To the best of our knowledge, this is the first time closed-loop control of an aerial robot with real-time flow field measurements has been demonstrated. 
Event cameras are chosen for their unique ability to deliver low-latency, motion-blur-free data under difficult lighting conditions, without the complexity, cost, and bandwidth limitations of traditional high-speed imaging systems with pulsed lasersheet illumination. 
\rebuttal{
Following standard practice in experimental fluid dynamics, velocimetry is performed using an external camera rather than an onboard sensor, which simplifies the measurement process and avoids measurement-induced flow perturbations.
}

\subsection*{Contribution}
\label{ssec:contribution}
We demonstrate the effectiveness of event-based vision for low-latency disturbance estimation of a drone flying inside a narrow pipe. Our scientific contributions are:
\begin{itemize}
    \item the first event-based smoke velocimetry method; it has a sub-millisecond processing latency, \rebuttal{runs on embedded computers,} and achieves a mean error of \unit[0.35]{m/s} w.r.t. offline estimates obtained with the SoTA software PIVLab~\cite{thielicke2014pivlab, thielicke2021pivlab, thielicke2022motionblur},
    \item the first disturbance estimator for quadrotors based on real-time flow field measurements; it estimates a horizontal and vertical disturbance force along with a roll-torque disturbance, and
    \item a neural controller trained via reinforcement learning that implicitly learns how to use the real-time estimates to counteract aerodynamic disturbances; this leads to improved control performance with a \unit[29]{\%} reduction in hovering position deviation and a \unit[71]{\%} reduction in overshoot during lateral position changes.
\end{itemize}
In addition, our engineering contributions are:
\begin{itemize}
    \item the development of the first monocular event-based motion-capture system; it has a sub-millisecond processing latency, \rebuttal{runs on embedded computers,} and achieves millimeter accuracy while only using small IR-LEDs as active markers, and
    \item an autotuning approach for event cameras using particle-swarm optimization~\cite{kennedy1995particleswarmoptimization} to optimize all camera biases.
\end{itemize}

\section{Related Work}
\label{sec:related_work}
This work draws inspiration from the fields of fluid mechanics, event-based vision, as well as machine learning for control. 
In this section, we present a summary of the related works in each of the fields, first looking at velocimetry methods and their applications to active flow control. 
Then we look at quadrotor flight in narrow pipes, and finally, a brief overview of monocular pose estimation is presented. 

\subsection{Image Velocimetry}
\label{ssec:image_velocimetry}
In particle image velocimetry (PIV), particles suspended in a flow are illuminated by a laser sheet and then tracked in order to obtain a time-resolved velocity field of the flow. Since its original development in 1984~\cite{adrian1984piv}, the method has been established as the standard method to measure flows~\cite{westerwheel1997piv, adrian2011piv}. Most commonly, one or more high-speed cameras are used for imaging and the particles are illuminated with a (pulsed) laser sheet.
Extensions enabling tracking of three-dimensional flows exist~\cite{schanz2016shakethebox,elsinga2006tomopiv}.

\subsubsection{Event-based Particle Image Velocimetry}
\label{sssec:event_based_particle_image_velocimetry}
To reduce the cost and complexity of the PIV setup and facilitate real-time data processing, the concept of event-based particle image velocimetry (EBIV) has been proposed in~\cite{drazen2011towardrealtimeparticletracking}, where the high-speed cameras are replaced with event cameras.
Subsequent works have refined the methodology and achieved stereo EBIV~\cite{wang2020stereoeventbasedpiv}, monocular EBIV with continuous laser illumination~\cite{willert2022ebiv}, and with pulsed laser illumination~\cite{willert2023pulsedebiv}, as well as real-time micro-particle tracking~\cite{rusch2023trackaer}.

\subsubsection{Smoke Image Velocimetry}
\label{sssec:smoke_image_velocimetry}
Classic PIV requires seeding mechanisms to introduce neutrally buoyant tracer particles into the flow. 
Smoke image velocimetry (SIV) has been proposed as a simpler, alternative approach to analyze airflow by using smoke as a tracer~\cite{mikheev2016siv}. 
The velocity field is then extracted by estimating the optical flow with template matching and subsequent subpixel refinement. 
This method has been used to analyze properties of a boundary layer~\cite{mikheev2017siv} and to extract turbulence parameters~\cite{wang2021siv}. However, SIV is generally not widely used in fluid mechanics and has not been realized with event cameras.

\subsubsection{Event-Based Optical Flow}
\label{sssec:event_based_optical_flow}
On a high level, event-based smoke velocimetry (EBSV) is very similar to event-based optical flow estimation.
Over the years, many optimization-based~\cite{bardow2016simultanousopticalflow, gallago2018contrastmaximization,shiba2024secrets} as well as learning-based methods~\cite{zhu20108evflownet,zhu20108evflownet, hagenaars2021selfsupervisedlearning} have been developed, demonstrating impressive capabilities on datasets like MVSEC~\cite{zhu2018mvsec} and DSEC~\cite{gehrig21dsec}, where an event camera is moved through a potentially dynamic scene.
Yet, there are subtle differences between the settings commonly studied and EBSV, making the latter a challenge of its own. 
First, smoke typically has a soft, edgeless, blob-like structure; and second, brightness constancy is violated by smoke moving in and out of the lightsheet, scattering light and potentially lighting up smoke outside the plane of the lightsheet.

\subsubsection{Active Flow Control}
\label{sssec:active_flow_control}
Most of the research in fluid dynamics is purely observational, with the notable exception of active flow control (AFC). 
In a nutshell, the goal of AFC is to use real-time measurements of the flow field in combination with an actuator to modify the flow towards a desired state~\cite{arnoult2024experimentalclosedloopflowseparationcontrol}. 
AFC can be combined with PIV for measuring the flow and providing real-time feedback to the controller, and in~\cite{varon2019adaptivecontrol}, a chaotic bimodal wake is stabilized into periodic oscillations with AFC. 
More recently, deep-learning-based AFC methods have been developed in the context of airflow separation~\cite{shimomura2020closedloopflowseparation} and drag reduction~\cite{lee2023turbulencecontrol}; however, these approaches typically rely on pressure sensing to measure the flow state or are open-loop~\cite{Ghraieb2021singlestepdeeprl}.

Note that for the works presented above, the goal state is defined in terms of the flow, not the actuator. For the task of flying a quadrotor in a narrow pipe, however, the goal state is defined for the quadrotor, not the flow itself. A somewhat similar scenario is studied with a toy problem in~\cite{roberts2012control}, where the goal is to control a hydrodynamic cartpole. Classic PIV is used to give feedback about the flow state to the controller. 
To the best of our knowledge, no prior works on active flow control in the context of real-world robotics exist.

\subsection{Quadrotor Flight in Confined Spaces}
\label{ssec:quadrotor_flight_in_confined_spaces}
Most relevant to this work are two recent prior works on quadrotor flight in confined spaces. In~\cite{wang2025autonomousflighttunnel}, the authors demonstrate fully autonomous flight inside narrow tunnels with fully onboard state estimation and control. 
The drone is capable of flying through square-shaped and cylindrical tunnels oriented either horizontally, vertically, or at an angle.
All state estimation is performed onboard.
However, to avoid its own recirculation, the drone cannot hover and must continually fly forwards. 
The authors use CFD and experiments to determine the optimal forward speed to be around \unit[1]{m/s} to achieve safe pipe traversal. 
The drones are about \unit[0.4]{m} in size, with the tunnels being at least \unit[0.6]{m} in width and many meters in length.

Concurrent research~\cite{martin2025flyinginducts} demonstrates hovering flight in circular tunnels. 
Either an external motion capture system or onboard time-of-flight sensors are used to provide a state estimate to the controller.
The key idea of the work is to use a robotic arm to measure the disturbance force inside the pipe as a function of the position and then integrate this information into the controller. 
When comparing the control performance inside and outside the pipe, they find much larger position tracking errors, up to \unit[6]{cm}, when flying inside the pipe. 
The quadrotor used in~\cite{martin2025flyinginducts} is similar in size and weight to the vehicle used in this work. 
Furthermore, the pipe diameter is \unit[35]{cm}, which is very similar to our experimental setup with \unit[38]{cm} inner diameter, making results potentially directly comparable.

While both works demonstrate flying inside the pipe, their focus is on state estimation and a control strategy to minimize the effects of the aerodynamic disturbances.
They are not concerned with the nature of the flow that develops, nor do they incorporate any real-time feedback about the flow in their control strategy\textemdash a research gap addressed by our work.

\subsection{Monocular Motion-Capture}
\label{ssec:monocular_motion_capture}
As outlined in the introduction, state estimation inside the pipe is challenging, as off-the-shelf motion capture systems require the quadrotor to be in view of multiple cameras~\cite{vicontracker2016manual}, making deployment in narrow, confined spaces extremely difficult. Consequently, we need to rely on monocular pose estimation. The approaches can be categorized into three families:
\begin{enumerate}
    \item standard cameras with fiducial markers (e.g., Aruco), 
    \item standard cameras with active point markers (e.g., LEDs), 
    \item event cameras with active markers (e.g., blinking LEDs).
\end{enumerate}
Independent of the marker type, standard cameras are not well-suited for a low-latency monocular motion capture system, as the latency of the overall system is strongly limited by the frame rate of the camera.
In contrast to a frame-based camera, an event camera is inherently low-latency, as events are available with microsecond latency. A major challenge is to design algorithms in a way that they are able to make use of this high update rate.

The first work~\cite{censi2013activeled} using event cameras for localization uses blinking LEDs as active markers that can be identified based on their frequency. However, the pipeline is not able to estimate the location along the optical axis of the event camera. Furthermore, it has noise levels on the order of \unit[10]{cm}, and thus is unsuitable for closed-loop control inside a narrow pipe.

More recent monocular event-camera pose estimation pipelines~\cite{salah2022neuromorphicvisionbased, ebmer2024realtime6dof} rely on very large active markers, up to $\unit[60\times 60]{cm}$, to achieve accurate localization. The markers also contain multiple LEDs such that their identification is not just via frequency detection, but similar to fiducial markers. 

While our system relies on similar concepts as~\cite{censi2013activeled}, we achieve unprecedented millimeter-accuracy, millisecond-latency pose estimates inside the pipe due to accurate and robust marker detection. The overall performance is comparable to the multi-camera, commercial motion capture systems ubiquitous in mobile robotics~\cite{ducard2009fma, menolotto2020mocapreview}.
\section{EBSV Real-Time Disturbance Estimation}
\label{sec:event_based_real_time_flow_estimation}
To achieve real-time disturbance estimation with event-based smoke velocimetry (EBSV) for quadrotor flight, both a low-latency flow estimator and a vision-based disturbance estimator that translates the observed flow into a disturbance wrench (force + torque) are required. 
Following the methodology commonly used for image velocimetry in fluid mechanics, we observe the flow in a two-dimensional plane with the help of a lightsheet that is perpendicular to the longitudinal axis of the pipe, as illustrated in Fig.~\ref{fig:1}b. The physical flow is then estimated from the optical flow.
To avoid ambiguities, the term \emph{optical flow} is used for the quantity estimated based on images, and \emph{flow} refers to the physical flow inside the pipe.

\subsection{Preliminary Considerations}
\label{ssec:preliminary_considerations}
Due to the nature of a lightsheet system, only a two-dimensional image of the potentially three-dimensional flow inside the pipe is observed. 
Because we consider a long pipe with $L/D > 10$ (e.g., its diameter is much smaller than its length), the time-averaged longitudinal flow $\langle v_x\rangle$ will be practically zero~\cite{pope2000turbulentflows}. 
However, the instantaneous, local flow near the quadrotor can be non-zero longitudinally, e.g., {$v_x \neq 0$}.
Assuming the lightsheet has a non-zero thickness $d_x$, the flow velocity is $v_x$, and assuming that a smoke structure can be tracked if not more than half of it leaves the illuminated region, the timescale $t_t$ at which a smoke structure can be meaningfully tracked is {$t_t = 1/2~d_x/v_x$}. 
For a lightsheet thickness of \unit[2]{cm} and a local, longitudinal flow velocity around \unit[1]{m/s}, this results in a tracking timescale of $t_t=$\unit[10]{ms}.

\begin{figure*}
    \centering
    \includegraphics{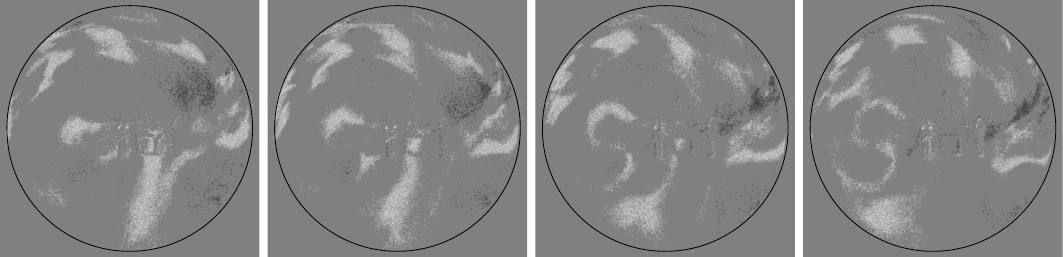}
    \caption{
    \rebuttal{
    Event frames of the turbulent airflow inside the pipe. The frames are spaced \unit[10]{ms} apart and polarity-integrated over a \unit[6]{ms} window. Characteristic smoke structures change appearance on very short timescales, necessitating a high velocimetry update rate and the use of larger spatial patches rather than dense, pixel-level flow estimates.
    }
    }
    \label{fig:event_flow}
\end{figure*}

This back-of-the-envelope calculation clearly shows three aspects: 
First, the flow estimation algorithm must run at very high update rates that are much faster than \unit[100]{Hz} in order to be able to track features, since they can disappear as quickly as \unit[10]{ms}. 
Second, obtaining per-pixel optical flow estimates is meaningless, as the observable smoke structures undergo appearance changes constantly. We need to evaluate the flow on larger scales. 
Third, the assumption of brightness constancy fundamental to many event-based methods is violated, as smoke structures can enter/exit the lightsheet. The scattered light can also illuminate the whole pipe, as it will contain some smoke.
\rebuttal{
Figure~\ref{fig:event_flow} exemplarily shows four event frames spaced \unit[10]{ms} apart to illustrate the points above.
}

\subsection{SSD Template Matching with Quadratic Refinement}
\label{ssec:ssd_template_matching}
Given the considerations presented above, we chose to follow the standard template matching approach from smoke image velocimetry~\cite{mikheev2017siv}, rather than a more modern unsupervised network-based optical flow.

A template-matching approach allows for highly parallel processing and direct control over the computational complexity by setting the grid size, patch size (referred to as interrogation window size in fluid dynamics literature), and maximum displacement. 
Furthermore, reliance on event-frames and template matching is robust to changes in overall brightness.
A quantitative comparison with a state-of-the-art, unsupervised learning approach and contrast maximization is presented in Sec.~\ref{ssec:optical_flow_estimation}. 

To achieve real-time performance, the algorithm design is tightly coupled with the C++/CUDA implementation. First, we convert the event-stream into frames and use a {$2 \times 2$} binning to spatially downsample the data. Formally, let
\begin{equation}
\mathcal{E} = \left\{e_i = \left(x_i, y_i, t_i, p_i\right),~p_i\in\{-1, 1\},~i = 1\hdots N\,\right\},
\end{equation}
denote the set of events, then the intensity value at pixel $(x,y)$ of the event frame $I_k$ at time $t = k\Delta t$ is given by
\begin{equation}
I_k(x,y) = \sum_{i = 1}^N p_i \cdot \begin{cases}
    1 & \text{if } \left\lfloor x_i/2 \right\rfloor = x ~\land~ \left\lfloor y_i/2 \right\rfloor = y \\
    & \land~ (k-1) \Delta t \leq t_i < k \Delta t \\
    0 & \text{otherwise}
\end{cases}
\end{equation} 
Conversion to event-frames is done on the CPU, as it can be implemented efficiently in linear time w.r.t. the length of the event stream, but for the subsequent steps the resulting event-frames are transferred to the GPU. 

\subsubsection{Template Matching}
When dealing with low-contrast smoke structures, the events are sparse because they are triggered probabilistically by the structures.
To increase the number of events, we use overlapped event time-windows of length $n=3$, \rebuttal{resulting in event-frames as shown in Fig.~\ref{fig:event_flow}}.
To avoid false minima in template matching, the image $I_k$ is blurred with a Gaussian kernel $G$ of size $7\times 7$ and a standard deviation of $\sigma_\text{blur}=1.75$:
\begin{equation} I_k^b = \sum_{j=k-n+1}^k I_j * G \end{equation}
The blurred accumulated image $I_k^b$ is then subdivided into $P \times P$ patches of size $w \times w$, and template matching with the previous image $I_{k-1}^b$ is performed for each patch. 
Consider a patch whose top-left corner is located at ${(x_0, y_0)}$. We use a normalized SSD (sum-of-squared-differences) measure to obtain a cost function $J(u, v)$ as a function of the displacement vector $(u,v)$ between the patches. Formally, let $D(u, v)$ denote the SSD for a patch for a given {$(u,v)$}
\[ 
D_k(u,v)\!=\!\!\!\sum_{x=x_0}^{x_0+w}\sum_{y=y_0}^{y_0+w}\!\left[I_k^b(x, y)\!-\!I_{k-1}^b(x\!+\!u\!-\!\frac{w}{2}, y\!+\!v\!-\!\frac{w}{2})\right]^2\!\!\!,
\]
and let $R_{k}(u,v)$ denote the autocorrelation
\[
R_k(u,v) =\sum_{x=x_0}^{x_0 +w}\sum_{y=y_0}^{y_0+w} I_k^b(x+u-\frac{w}{2},y+ + v + \frac{w}{2})\;.
\]
Then, the cost function is defined by
\begin{equation}
J_k(u,v) = \frac{D_k(u+w/2,v+w/2)}{\sqrt{R_k(w/2, w/2) R_{k-1}(u+w/2,v+w/2)}}
\end{equation}
The optical flow $\mathbf{u}$ is then given as the solution to the minimization 
\begin{equation}
\mathbf{u}^*_k = [u_k^*~~v_k^*]^\top = \argmin_{u, v} J_k(u, v)
\label{eq:ssd_optimization}
\end{equation}
We solve (\ref{eq:ssd_optimization}) by evaluating it for the grid $u\in [-u_\text{max},u_\text{max}]$ and  $v\in [-v_\text{max},v_\text{max}]$ in parallel on the GPU and use a maximal displacement $u_\text{max}=v_\text{max}=8$.

\subsubsection{Quadratic Refinement}
The above approach can only yield pixel-accurate results, but especially at high update rates the relative displacement between two consecutive frames can be small, and hence sub-pixel accuracy is desired. This is achieved by fitting a quadratic surface to $J_k(u,v)$ through a {$3 \times 3$} neighborhood centered at $(u^*, v^*)$~\cite{mikheev2016siv}, i.e., we find the ordinary least-squares solution $\mathbf a^*$ to the problem
\begin{align*} 
&a_0 + a_1 u + a_2 v + a_3 uv + a_4u^2 + a_5v^2 = J_k(u,v) \\
&\text{with } u = u^* - 1 \hdots u^* + 1,~~ v = v^* - 1 \hdots v^* + 1\;,
\end{align*}
and then find the minimum of the quadratic surface as the solution to the problem $\mathbf A \mathbf u_\text{subpx} = \mathbf b$ with
\[ 
\mathbf A = \begin{bmatrix} 2 a^*_4 & a^*_3 \\ a^*_3 & 2 a^*_5 \end{bmatrix},\quad \mathbf b = \begin{bmatrix}
    -a^*_1 \\ -a^*_2
\end{bmatrix}
\]
The curvature of the quadratic surface is related to $\text{det }\mathbf A$, and we discard all flow estimates with {$\text{det }\mathbf A \leq 0$} (maximum, not minimum) and {$||\mathbf{u}_\text{subpx}||_\infty > 1$} (minimum outside region). Additionally, we empirically find that $\sqrt{\text{det }\mathbf A}$ is a good estimate of the confidence in the correctness of the flow estimate of a patch.

The final output of our optical flow estimator is a 3-channel image $\varphi \in \mathbb{R}^{P \times P \times 3}$. Each pixel {$i,j$} represents the optical flow %
\[ \varphi(i,j) = \begin{bmatrix}u^* + u_\text{subpx} & v^* + v_\text{subpx} & \sqrt{\text{det } \mathbf A} \end{bmatrix}^\top \]
for a $w \times w$ pixel patch at {$(x_0, y_0) = (i_0 + i\Delta, j_0 + j \Delta)$} in the original image. We chose ${i_0 = j_0 = u_\text{max}}$, step size {$\Delta = 24$}, and window size $w = 32$. If a flow estimate got discarded during the refinement step, the vector is set to all zeros.

\subsubsection{Resolution and Latency} 
\label{ssec:flow_latency}
In the plane of the lightsheet, we get a downsampled resolution of \unit[0.8]{px/mm}. 
Assuming a measurement accuracy of \unit[0.5]{px} and running the flow estimator with $\Delta t=\unit[2]{ms}$ (\unit[500]{Hz}), we can thus minimally measure flow speeds of \unit[0.31]{m/s}.
With the chosen maximal per-frame displacement of 8 pixels, the maximum measureable (axis-aligned) speed is \unit[5]{m/s}, which corresponds to the induced velocity at hover of the quadrotor.

\begin{table}[b!]
    \vspace*{-9pt}
    \caption{Runtimes of the EBSV Algorithm}
    \label{tab:runtimes_eventflow}
    \vspace*{-3pt}
    \setlength{\tabcolsep}{2pt}
\begin{tabularx}{1\linewidth}{lC|lC}
    \toprule
    Embedded & Runtime [ms] & Workstation (CPU + GPU) & Runtime [ms] \\
    \midrule
     Jetson Orin NX  & 4.98 &  i9 8950HK + P3200    & 1.58  \\
     Jetson Thor     & 1.09 &  i9 12900K + RTX 3090 & 0.44  \\
                     &      &  9950X3D + RTX 5090   & 0.33  \\
    \bottomrule
\end{tabularx}

\end{table}

Stacking the event frames, blurring the stacked image, and computing the cost function are implemented on the GPU using Nvidia's NPPPlus SDK~\cite{nvidia2024npplus} and accelerated with graph tracing. 
\rebuttal{
We evaluate the performance of our EBSV-algorithm on both embedded systems and general-purpose computers. The runtimes for $P=11$ (121 patches) and all other parameters as given above are summarized in Tab.~\ref{tab:runtimes_eventflow}. The credit-card sized Orin NX is already capable of achieving \unit[200]{Hz} and more powerful embedded systems such as the Nvidia Jetson Thor reach close to \unit[1]{kHz}. Workstations can run the flow estimator in real-time at over \unit[500]{Hz}, with a modern GPU yielding sub-millisecond latency.
}

\subsection{Wrench Estimation}
\label{ssec:wrench_estimation}
\begin{figure}
    \centering
    \includegraphics{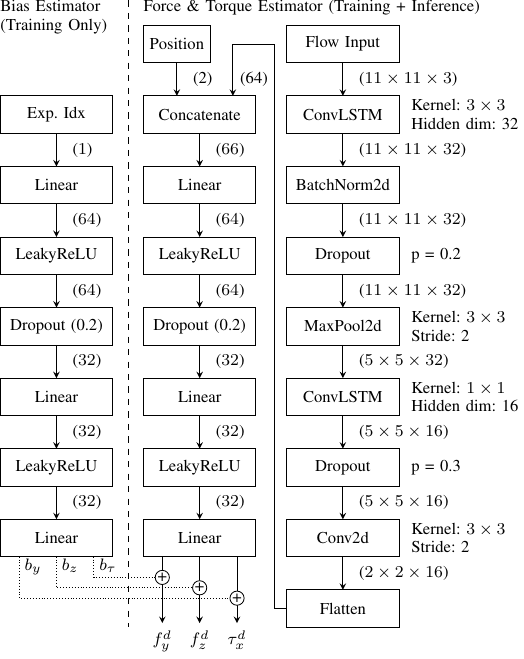}
    \vspace*{-12pt}
    \caption{Our network architecture used to estimate the aerodynamic disturbances. The ConvLSTM-based flow encoder outputs a latent representation of the flow configuration. This is concatenated with the position of the quadrotor and fed into an MLP to estimate the aerodynamic disturbances. \rebuttal{During training we also train with an additive bias which is only a function of the experiment index and accounts for experimental biases, e.g. battery placement.}}
    \label{fig:network_architecture}
    \vspace*{-12pt}
\end{figure}
To estimate aerodynamic disturbances acting on the quadrotor, we design a neural network architecture that leverages the sparse optical flow measurements $\varphi$ obtained through our event-based smoke velocimetry pipeline. 
The network is visualized in Fig.~\ref{fig:network_architecture} and has two inputs. The first is the quadrotor's $y$ and $z$ position within the pipe's cross-section to provide spatial context.
The second input is the optical flow $\varphi$. Note that including the confidence value as a third channel allows the network to assign greater weight to more reliable optical flow inputs.

The flow input is processed using a convolutional Long Short-Term Memory (ConvLSTM) network. 
This module enables the network to capture both spatial and temporal patterns within the flow field and smooth out noisy flow measurement vectors. 
\rebuttal{We opted for the ConvLSTM architecture over more recent video-transformers~\cite{arnab2021vivit} because their spatio-temporal attention and frame batching makes them less suited for sequential, real-time applications.}
The output of the ConvLSTM is a compact latent representation.
This latent vector is then concatenated with the position input and passed to a two-layer perceptron (MLP), which outputs estimates of the horizontal disturbance force $f_y^d$, vertical disturbance force $f_z^d$, and roll torque disturbance $\tau_x^d$ acting on the vehicle.

A significant challenge in estimating aerodynamic disturbances is the presence of experimental biases. 
Such biases arise from physical asymmetries in the quadrotor (e.g., unbalanced mass distribution due to off-center battery placement). 
While these biases are approximately constant within a single experiment, they can vary across different trials and corrupt the learned disturbance signal. 

To address this, we introduce a parallel bias estimation module, implemented as a lightweight MLP, which is trained jointly with the primary disturbance network \rebuttal{as shown in Fig.~\ref{fig:network_architecture}.}
This is only used during training and discarded afterwards. 
The bias estimation module only has access to a single input, the index of the experiment (simply an increasing sequence number), and thus can only estimate an additive, per-experiment bias.
The total loss function $\mathcal{L}$ used for training combines two terms: a mean squared error (MSE) loss between the bias-corrected, predicted disturbance $\hat d + \hat b$, and the ground-truth disturbance $d_\text{gt}$; and an L2 regularization term on the estimated bias.
\begin{equation}
\mathcal{L} = \text{MSE}(\hat{d} + \hat{b}, d_{\text{gt}}) + \lambda \|\hat{b}\|^2
\end{equation}
The regularization ensures that the bias network only compensates when necessary, allowing the model to disentangle systematic biases from dynamic flow-induced effects.
\rebuttal{
Compared to bias-mitigation strategies that operate by enforcing symmetry in the estimated wrenches, our approach has the advantage to be applicable to arbitrarily shaped pipes and other drones as it makes no assumptions on the cause of the bias.
}

\section{Event-Based Monocular Pose Estimation}
\label{sec:event_based_monocular_pose_estimation}

\begin{figure*}
    \centering
    \includegraphics[width=1\linewidth]{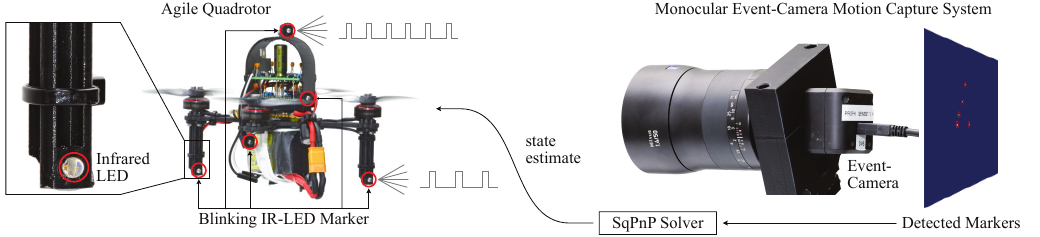}
    \caption{Overview of our monocular event-camera motion capture system: the quadrotor is equipped with $N>=4$ infrared LED markers that blink at different frequencies. A single, calibrated event-camera is used to detect all markers and estimate the pose of the object by solving the perspective-n-points problem (PnP) with SqPnP~\cite{terzakis2020sqpnp}.}
    \label{fig:event_vicon}
\end{figure*}

Having addressed the topic of aerodynamic disturbance estimation in the previous section, we now focus on the next challenge: state estimation. 
A monocular event-based motion-capture system with active markers is ideally suited for this task.
Figure~\ref{fig:event_vicon} shows the developed millisecond-latency, millimeter-accuracy motion capture system, which takes inspiration from~\cite{censi2013activeled}.
We first give an overview of the whole system and then describe the two key aspects enabling robust, accurate, and low-latency pose estimation: the novel SDTV (Signed Delta-Time Volume) event representation enabling a processing latency of $\unit[273]{\mu s}$ on a laptop and our particle-swarm optimization, which optimally sets the contrast thresholds, biases, and filter parameters of the event camera.
Details on the electrical design of the active LED markers and the pose estimation accuracy can be found in App.~\ref{sec:blinking_led_circuit} and \ref{sec:pose_estimation_noise}.

\subsection{Overview}
In contrast to multi-camera motion-capture systems that rely on triangulation, our monocular system must solve the PnP (perspective-n-point problem)~\cite{zisserman2004multipleview} to obtain the pose of the object in the camera frame. 
For a unique solution, at least four 3D~$\leftrightarrow$~2D correspondences must be known~\cite{gao2003complete}. 
In the PnP setting, it is not enough to detect the markers; the markers must also be uniquely identified to establish the 3D~$\leftrightarrow$~2D correspondences.  Note that the locations of the markers on the tracking object are assumed to be known, for example from a CAD model.
By using blinking LEDs as active markers, they can be easily detected by the event-camera. To uniquely identify each marker, the LEDs blink at different, known frequencies. 
The blinking frequency for each detected marker is then measured from the event stream to associate the detection in the image with an LED marker.
The high temporal resolution of event cameras makes it possible to use very fast blinking frequencies and obtain a low-latency system, and we use between {$f_\text{min}=\unit[1.7]{kHz}$} and {$f_\text{max}=\unit[2.9]{kHz}$}.

Once the active markers are detected and their identities extracted, a particle filter is used to track the centroids of each marker. Finally, the PnP problem is solved with SqPnP~\cite{terzakis2020sqpnp} to estimate the pose of the quadrotor. In agreement with~\cite{terzakis2020sqpnp}, we find that SqPnP is more accurate than EPnP~\cite{lepetit2009epnp} (see App.~\ref{sec:pose_estimation_noise}).

\subsection{Efficient Event Data Representation}
\label{ssec:efficient_event_data_representation}
A key element that enables our event-camera system to run at rates exceeding \unit[1]{kHz}, even on a laptop, is the novel event data representation, which has been optimized for computational performance. 

A single event, as supplied by the camera, is given as a four-tuple consisting of an $x$ and a $y$ coordinate (both \texttt{uint16\_t}), a polarity $p$ which is either -1 or 1 (\texttt{int8\_t}), and a timestamp $t$ (\texttt{uint64\_t}). Additionally, 3 bytes of padding are included for 16-byte alignment.
In the context of blinking LED detection, the representation must be efficient to process both in the time domain (frequency identification) and the spatial domain (clustering). Event-frames and time-surface representations are efficient for spatial operations~\cite{gallego2022eventbasedvision} but do not encode timestamps in a way that can be used for frequency identification. An event-volume/event-voxel representation would need very fine binning in the temporal axis for accurate frequency identification and thus would require far more storage than fits in the processor cache, adversely affecting runtime.

\subsubsection{Signed Delta-Time Volume (SDTV)}
\begin{figure}
    \centering
    \includegraphics{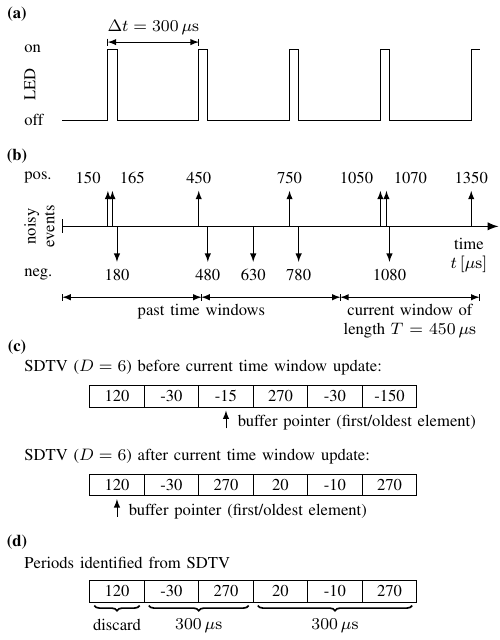}
    \caption{Illustration on the construction of the \emph{Signed Delta-Time Volume (SDTV)} from an event stream. 
    \textbf{a)}~The LED is blinking with a period of $\unit[300]{\mu s}$ and a duty cycle of \unit[10]{\%}. 
    \textbf{b)}~A single pixel of the event camera records a noisy signal of this blinking LED. False double events (e.g., at $t = \unit[150]{\mu s}, \unit[165]{\mu s}$) and spurious events (e.g., at $t = \unit[630]{\mu s}$) are included. 
    \textbf{c)}~Construction of the SDTV illustrated before and after processing the latest time window. 
    \textbf{d)}~Periods robustly identified from the SDTV by summing up absolute time differences between negative $\rightarrow$ positive transitions (the first positive value is included). All events until the first positive $\rightarrow$ negative transition are discarded.}
    \label{fig:sdtv}
    \vspace*{-6pt}
\end{figure}

To address the shortcomings of those widely used event representations, we propose a data representation that is ideally suited for the task of blinking LED detection: the \emph{signed delta-time volume (SDTV)}. 
It is a 3D volume of size {$W \times H \times D$}, where $W$, $H$, and $D$ are the sensor width, sensor height, and SDTV stack depth, respectively.
For each pixel, the time difference to the last event is stored, and the polarity of the event is encoded in the sign of this time difference. 
This is possible because time must be monotonically increasing, so we can repurpose the sign bit for polarity encoding. 
Consequently, the polarity information and the time information are available with a single load instruction to the CPU. 
The idea is illustrated in {Fig.~\ref{fig:sdtv}a,b} for a single pixel stack of the SDTV.

Because of the fast blinking of the LEDs, the time differences (in microseconds) between consecutive events are always within \texttt{int16\_t} range, making storage compact and cache-friendly. 
As most operations are done per pixel stack, the memory layout is such that the $D$-dimension is continuous. 
Similar to the other representations, converting an event stream to SDTV is linear in the number of events.
The signed delta-time volume is not computed per window of length $T$ but updated as a cyclic buffer, as shown in Fig.~\ref{fig:sdtv}c,d. 
This increases the robustness of the frequency detection for LEDs blinking at a lower frequency than $f_\text{max}$, since the number of LED periods available for frequency identification is independent of the frequency and constitutes a practically highly relevant improvement of~\cite{censi2013activeled}.

\subsubsection{Latency}
\label{ssec:vicon_latency}
\rebuttal{
To fully take advantage of the low latency of the event camera, the processing must fast. In Tab.~\ref{tab:runtimes_evmocap} we show the performance of our approach on different devices, ranging from embedded devices to desktop workstations. Notably, even on an old Raspberry Pi 4B we are able to run our motion-capture at \unit[300]{Hz} and on modern Nvidia boards, speeds above \unit[1]{kHz} are easily achieved. Consequently, we are able to truly leverage the low latency of the event camera and can also deploy the system on embedded hardware.
}

\rebuttal{
Looking at the breakdown of the overall runtime we note that the relative runtime of the SDTV-based LED detection and frequency estimation is much larger on the older ARM-cortex based systems (Raspberry Pi 4B and Jetson Orin NX). A possible explanation for this is the processor cache size which is \unit[1]{MB} on the Raspberry Pi 4B and \unit[4]{MB} on the Orin NX. At VGA resolution and $D=16$ the SDTV volume is \unit[9.4]{MB} in size and thus does not fit into the cache. Frequent cache misses negatively affect the runtime, a problem that the Thor and other CPUs do not have, as their L3-caches are large enough to fit the entire SDTV volume.
}

\begin{table}[bp]
    \vspace*{-6pt}
    \caption{Runtimes of the Event-based Pose-Estimation}
    \label{tab:runtimes_evmocap}
    \vspace*{-3pt}
    \renewcommand{\arraystretch}{1.3}
\setlength{\tabcolsep}{5pt}
\begin{tabularx}{\linewidth}{cl|C|ccc}
    \toprule
     & Device & \multicolumn{4}{c}{Runtimes [ms] (\% of total runtime)} \\
     & & Total & Detect LED & Filter & SqPnP \\
    \midrule
    \multirow{3}{*}{\rotatebox{90}{Embedded}} 
    & Raspberry Pi 4B   & 3.22 & 2.48 (77\%) & 0.40 (12\%) & 0.34 (11\%) \\
    & Jetson Orin NX    & 1.29 & 0.92 (71\%) & 0.23 (18\%) & 0.14 (11\%) \\
    & Jetson Thor       & 0.61 & 0.38 (62\%) & 0.14 (23\%) & 0.09 (14\%) \\
    \midrule
    \multirow{3}{*}{\rotatebox{90}{Computer}} 
    & Intel i9 8950HK   & 0.27 & 0.14 (52\%) & 0.07 (26\%) & 0.06 (22\%) \\
    & Intel i9 12900K   & 0.17 & 0.09 (53\%) & 0.05 (29\%) & 0.03 (18\%) \\
    & AMD 9950X3D       & 0.14 & 0.06 (43\%) & 0.04 (29\%) & 0.04 (28\%) \\
    \bottomrule
\end{tabularx}

\end{table}

Note that due to the slowest LED blinking only at ${f_\text{min}=\unit[1700]{Hz}}$, increasing the processing frequency beyond $f_\text{min}/2$ does not make sense, as at least one full LED cycle must be captured to identify the frequency. Thus, the overall worst-case latency in this work is considered $2/f_\text{min} + t_\text{proc}$, e.g. $\unit[1.45]{ms}$ for the laptop workstation.

\subsection{Automatic Event Camera Tuning}
\label{ssec:automatic_event_camera_tuning}
To maximize the robustness of the SDTV-based frequency identification, there is ideally exactly one positive and one negative event per LED transition.
Achieving this is a difficult task, as event-camera manufacturers provide ample tuning parameters and little guidance on how to optimally set them.
We propose to use particle-swarm optimization (PSO)~\cite{kennedy1995particleswarmoptimization}, with the event-camera in the loop, to jointly optimize all tuning parameters. 
In contrast to existing methods~\cite{kilmaghani2025autobiasing}, which focus on real-time bias tuning, our method focuses on the optimality of the final parameters.

Consider a scenario where the event-camera and an object with $K$ markers with known frequencies ${f_k,~k=1\hdots K}$ are statically placed. 
The pixel coordinates of each LED center are given by {$(x_k^c, y_k^c)$}. 
At a pixel $(x,y)$ illuminated by marker $k$, an event-stream of duration $t$ with events $n = 1\hdots N$ ideally contains $t \, f_k$ positive {$p = 1$} and negative {$p = -1$} events.
We define the per-pixel event ratio $\alpha_k(x,y)^+$ for positive and $\alpha_k(x,y)^-$ for negative events as
\begin{align}
    \alpha_k(x,y)^+ &= \frac{\sum_{n = 1}^N 1 + p_n(x,y)}{2 t \, f_k} - 1 \\
    \alpha_k(x,y)^- &= \frac{\sum_{n = 1}^N 1 - p_n(x,y)}{2 t \, f_k} - 1
\end{align}

\begin{table}[bt]
    \centering
    \caption{Default and Optimal Parameters for Gen3}
    \label{tab:biases}
    \begin{tabularx}{1\linewidth}{l|CCCCCc}
    \toprule
    Parameter & diff\_off & diff\_on & bias\_fo & bias\_hpf & bias\_pr & bias\_refr \\ \midrule
    Default   & 225 & 375 & 1725 & 1500 & 1500 & 1500 \\
    Optimized & 176 & 529 & 1665 & 1724 & 1768 & 1538 \\
    \bottomrule
    \end{tabularx}
\end{table}

Optimally, $\alpha$ is zero. Cases with $\alpha < 0$ correspond to missing events, and $\alpha > 0$ correspond to multiple events per transition. It is somewhat acceptable to have multiple events per transition, but missing transitions severely affect the accuracy of the frequency identification. Consequently, we define an asymmetric per-pixel cost function $J_\text{px}$
\begin{equation}
    J_\text{px}(\alpha) = \begin{cases}
        4 \alpha^2 & \alpha < 0 \\
        0 & 0 <= \alpha <= \alpha_0 \\
        (\alpha-\alpha_0)^2 & \alpha > \alpha_0
    \end{cases}
\end{equation}
where $\alpha_0 = 0.5$ introduces some slack into the optimization to allow a \unit[50]{\%} chance of triggering two events per transition. To further increase the robustness, we consider a patch of {$3 \times 3$} pixels around the center pixel for the optimization, leading to an overall objective function $J$
\begin{align}
    J = \sum_{k = 1}^{K} \sum_{x_k^c - 1}^{x_k^c + 1}  \sum_{y_k^c - 1}^{y_k^c + 1}  J_\text{px}\left(\alpha_k^+(x,y)\right) + J_\text{px}\left(\alpha_k^-(x,y)\right)
\end{align}
We employ PSO~\cite{kennedy1995particleswarmoptimization} with 100 particles to jointly optimize the biases $\theta$ of the event camera to minimize the above cost function $J$. We set $t = \unit[0.1]{s}$ and after 40 to 60 iterations (around \unit[10]{min} time), the algorithm converges with an overall cost of $J^* = 0$, i.e., no events are missed and not more than \unit[50]{\%} of the transitions trigger two events. The optimal parameters for a Prophesee Gen3 are reported in Tab.~\ref{tab:biases}.
\rebuttal{The very high contrast thresholds also effectively eliminate spurious events triggered by the moving smoke and strongly attenuate noise. Measurements show a noise event-rate of only \unit[0.7]{ev/ms}, compared to \unit[587]{ev/ms} with the blinking LEDs. The SDTV representation and subsequent processing fully prevent any spurious marker detections due to noise.}

\section{Disturbance-Aware Quadrotor Controller}
\label{sec:disturbance_aware_quadrotor_controller}
With both disturbance estimation and state estimation addressed, the last component of our system is the learning-based controller which takes the real-time disturbance measurements into account.
Following the demonstrated success of reinforcement learning (RL) in enabling robust quadrotor flight in challenging real-world conditions~\cite{kaufmann23champion, song2023reaching}, we choose an RL-based approach in this work.
\rebuttal{
The RL-controller outputs collective thrust and body-rate (CTBR) commands~\cite{kaufmann2022benchmark}, which are tracked by a low-level PID controller onboard the drone. During deployment, the integral term of the low-level controller compensates for static torque biases, effectively masking them from the RL-controller.
}

\subsection{LSTM-PPO Policy Architecture}
\label{sssec:lstm_ppo_policy_architecture}
We adopt a model-free policy gradient method based on Proximal Policy Optimization (PPO)~\cite{schulman2017proximal}, owing to its widespread use and demonstrated success. 
While conventional quadrotor RL controllers are typically implemented as multilayer perceptrons (MLPs), such memoryless architectures cannot account for the temporal nature of turbulence. 
To address this, we use LSTM-PPO, a variant of PPO that incorporates a recurrent neural network in the policy~\cite{hausknecht2015deep}, allowing the controller to implicitly learn temporal dependencies.
The observation of the policy is defined as
\[ \mathbf o = \begin{bmatrix}
\mathbf{\hat p} &
\mathbf{\hat R}_{12}(:) &
\mathbf{p}_\text{target} & 
\mathbf{a}_{t-1} & 
\hat f_y^d & \hat f_z^d & \hat \tau_x^d & \zeta 
\end{bmatrix}^\top \in \mathbb{R}^{19} \]
where ${\mathbf{\hat p} \in \mathbb{R}^3}$ is the estimated quadrotor position, ${\mathbf{\hat R}_{12}(:) \in \mathbb{R}^6}$ denotes the vectorized first two columns of the rotation matrix representing estimated orientation, and ${\mathbf{p}_\text{target} \in \mathbb{R}^3}$ is the goal position. The term ${\mathbf{a}_{t-1} \in \mathbb{R}^3}$ denotes the action applied in the previous timestep. The estimated disturbance wrench is given by ${[\hat{f}_y^d, \hat{f}_z^d, \hat{\tau}_x^d] \in \mathbb{R}^3}$, and $\zeta \in \{0, 1\}$ is a binary indicator flag that signals whether valid disturbance measurements are available at the current timestep (e.g., is smoke currently injected into the pipe). This mechanism enables the policy to learn how to handle both nominal conditions (without flow sensing) and flow-aware control when the estimator is active.

We employ a privileged critic with access to the full system state, including velocities, angular rates, and ground-truth disturbance signals. Specifically, the critic observation is given by the concatenation of the noisy actor observation with ground-truth values:
\[
\mathbf{o}_\text{critic} = 
\begin{bmatrix}
\mathbf{o} & \mathbf{p} & \mathbf{R}_{12}(:) & \mathbf{v} & \boldsymbol{\omega} & f_y^d & f_z^d & \tau_x^d
\end{bmatrix}^\top 
\in \mathbb{R}^{38},
\]
where $\mathbf{v} \in \mathbb{R}^3$ and $\boldsymbol{\omega} \in \mathbb{R}^3$ denote the linear and angular velocities.

\subsection{Training in Simulation}
\label{ssec:training_in_simulation}
We leverage the \emph{Agilicious}~\cite{foehn2022agilicious} simulation framework to model quadrotor dynamics. A key challenge is the simulation of unsteady aerodynamic disturbances that reflect realistic pipe recirculation. 
We model the total disturbance as a superposition of a quasi-steady component capturing effects of a fully developed recirculation flow, and a time-varying stochastic component. 
The steady component is generated using a reduced version of our learned disturbance estimator (Fig.~\ref{fig:network_architecture}) which uses only the positional input. 
During training, this lightweight model provides a position-dependent mean disturbance profile.

The unsteady, turbulent component is modeled by analyzing real-world data. First, we subtract the mean, position-dependent disturbance from measured wrench data to isolate the residual. 
The power spectral density (PSD) of this residual is then estimated and used to fit an autoregressive (IIR) filter via the Yule-Walker method. 
During simulation, temporally correlated noise is generated by filtering Gaussian white noise through this IIR model, reproducing the correct spectral characteristics of the real-world disturbances.
\rebuttal{
To validate our simulation approach in the time-domain, we treat the real-world and simulated disturbances as samples from probability distributions and compute the Wasserstein distance $W_1$ between them. For the force disturbances along the $y$ and $z$ axes, we obtain distances of \unit[7.3]{mN} and \unit[11.6]{mN}, respectively. For the roll-torque disturbances, the distance is \unit[0.56]{mNm}, indicating close agreement between simulation and experiment.
}

In addition, we simulate realistic sensor noise in the pose measurements by analyzing the measurement accuracy of the monocular motion-capture system. Again, a PSD is estimated and used to generate temporally correlated observation noise, which is added to the ground-truth position and rotation observations to generate $\mathbf{\hat p}$ and $\mathbf{\hat R}$.

\subsection{Training Details}
\label{ssec:training_details}
We train the policy using 96 environments in parallel. In each episode, the quadrotor is initialized at a random position within the pipe, with a randomized target goal location. 
The reward function $r_t$ at time $t$ consists of four main components: a positional reward $r_\text{pos}$ to fly to the target location, a velocity reward $r_\text{vel}$ encouraging progress towards the destination, and two smoothness rewards $r_\omega$ and $r_\text{cmd}$:
\begin{equation}
r_t = r_\text{pos} + r_\text{vel} - r_\omega - r_\text{cmd},
\end{equation}
where
\begin{align*}
r_\text{pos} &= \lambda_1 \exp\left(-\tfrac{1}{2} (\mathbf{p} - \mathbf{p}_\text{target})^\top \Sigma_p^{-1} (\mathbf{p} - \mathbf{p}_\text{target}) \right), \\
r_\text{vel}  &= \lambda_2 \cdot \mathbf{v}^\top (\mathbf{p}_\text{target} - \mathbf p), \\
r_\omega &= \lambda_3\| \boldsymbol{\omega} \|^2, \\
r_\text{cmd} &= \lambda_4 \| \mathbf{u}_t - \mathbf{u}_{t-1} \|^2.
\end{align*}

An episode is considered successful if the quadrotor reaches the target within a threshold ($||\mathbf{p} - \mathbf{p}_\text{target}|| < \unit[0.01]{m}$) and stabilizes with low translational and rotational velocities ($||\mathbf{v}|| < \unit[0.1]{m/s}$, $||\boldsymbol{\omega}|| < \unit[0.1]{rad/s}$).

A simple curriculum is employed: aerodynamic disturbances are gradually introduced after 500 episodes and reach full magnitude by episode 2000. This easing-in improves convergence and enables the policy to first learn nominal flight behaviors before dealing with disturbance rejection. In total, we train for 4000 episodes (1e8 environment interactions), which takes \unit[2]{h} on a desktop workstation.
\section{Experimental Setup}
\label{sec:experimental_setup}

To experimentally verify the proposed approach for real-time flow estimation and closed-loop control of a drone, a suitable setup (illustrated in Fig.~\ref{fig:1}b) is needed.

To approximate the flow states in an infinite tube, the influence of the open ends on the flow should be minimal, and consequently, we opt for a pipe with {$L/D \gg 1$}~\cite{pope2000turbulentflows}. Specifically, a \unit[5]{m} long PVC pipe with \unit[38]{cm} inner diameter is used. To avoid internal reflections, the pipe is coated with anti-reflective black paint, typically used to paint the inside of telescope tubes. 
In the middle, 18 smoke injectors are installed. These nozzles feature a fairly large diameter of \unit[1]{cm} to minimize the flow velocity of the smoke and hence minimize the disturbance caused by injecting smoke into the flow.
Smoke is supplied by three off-the-shelf smoke machines, and we use fast-dissipating fog fluid to prevent the pipe from filling with smoke. 
In total, the three smoke machines can supply smoke for about \unit[20]{s} before heating up again for about \unit[60]{s}.
Directly next to the smoke injectors, the LED lightsheet is located. 
We build this by installing an LED strip in a recess outside of the tube with a small slit-aperture. While not as narrow as laser sheets, the LEDs provide constant illumination unlike pulsed like laser sheets.

\begin{table}[tb]
    \centering
    \caption{Specifcations of the Quadrotor}
\label{tab:kolibri_specs}
\begin{minipage}[c]{0.6\linewidth}
    \begin{tabularx}{1\linewidth}{X>$l<$|r>$l<$}
        \toprule
        Parameter & & Value & \text{Unit} \\ \midrule
        Mass & m & 0.21 & \unit{kg} \\
        Inertia & J_{xx} & 0.14 & \unit{g m^2} \\
        & J_{yy} & 0.17 & \unit{g m^2} \\
        & J_{zz} & 0.21 & \unit{g m^2} \\
        Propeller Radius & r_p & 3.81 & \unit{cm} \\
        Quadrotor Size & l & 19.43 & \unit{cm} \\
        Max Total Thrust & & 14 & \unit{N} \\
        \bottomrule
    \end{tabularx}
\end{minipage}
\begin{minipage}[c]{0.385\linewidth}
    \flushright
    \begin{tikzpicture}[>=stealth, scale=1]
        \def\l{0.85}
        \def\r{0.5}
        \def\sqrt2{1.414}
        \fill [fill=lightgray] (-\l, -\l) circle (\r); 
        \fill [fill=lightgray] (-\l, \l) circle (\r); 
        \fill [fill=lightgray] (\l, -\l) circle (\r); 
        \fill [fill=lightgray] (\l, \l) circle (\r); 
        \draw [lightgray, line width = 5pt] (\l, \l) -- (-\l, -\l);
        \draw [lightgray, line width = 5pt] (\l, -\l) -- (-\l, \l);
        \draw [Bar->] (\l,\l) -- node [above, midway] {$r_p$} (\l + \r,\l);
        \coordinate (a) at (-\l - \r / \sqrt2 , \l + \r / \sqrt2);
        \coordinate (b) at (\l + \r / \sqrt2, -\l - \r / \sqrt2);
        \draw [<->] ([xshift=-0.4cm, yshift=-0.4cm]a) -- node [pos = 0.4, rotate = -45, above] {$l$} ([xshift=-0.4cm, yshift=-0.4cm]b);
        \draw (a) -- ([shift={(-0.45cm, -0.45cm)}]a);
        \draw (b) -- ([shift={(-0.45cm, -0.45cm)}]b);
    \end{tikzpicture}
\end{minipage} 
    \vspace*{-12pt}
\end{table}

To perform the experiments, a small and lightweight quadrotor is equipped with the active LED markers detailed in Sec.~\ref{sec:blinking_led_circuit}. 
The specifications of the vehicle are summarized in Tab.~\ref{tab:kolibri_specs}. 
The quadrotor's tip-to-tip length of \unit[19.5]{cm} corresponds to half the pipe diameter. 
With a thrust-to-weight ratio of about~7, the quadrotor has enough thrust to avoid saturation of any motor even in challenging scenarios. 

For the monocular event-based state estimation system, a Prophesee Gen3 camera with a Zeiss Milvus \unit[50]{mm} f/1.4 lens is used. To ensure that the imaging system has sufficient depth of field, the lens is set to an aperture of f/5.6. 
For pose estimation, we assume that the event camera is calibrated, and we follow the approach proposed by~\cite{muglikar2021calibrate}, where the calibration is performed by first converting the event stream into event frames and then calibrating those using Kalibr~\cite{rehder2016extending}. 
The calibration also estimates the lens distortion, and in this work, we rely on a double-sphere distortion model because of its combination of accuracy and computational efficiency~\cite{usenko2018doublesphere}.

As shown in Fig.~\ref{fig:1}b, the event camera responsible for flow velocimetry is located on the opposite side of the pipe. We use a Prophesee Gen4 with a Zeiss Classic \unit[25]{mm} f/2 lens to match the field of view of both event cameras and install an IR-blocking filter to avoid seeing the blinking IR LEDs.

\begin{figure*}
    \centering
    \includegraphics{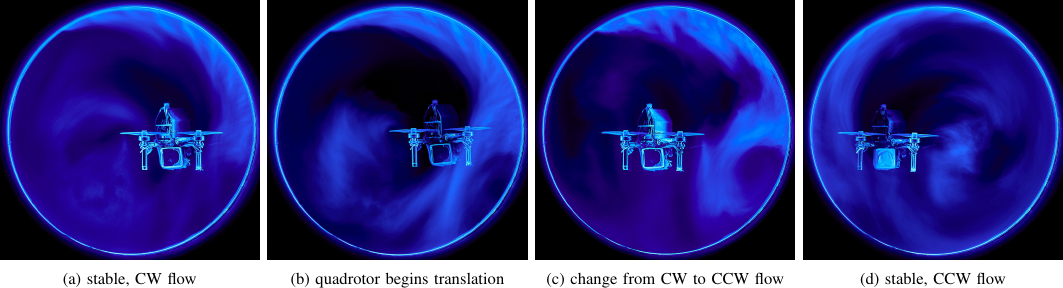}
    \caption{Flow configurations during a lateral translation across the pipe. \textbf{a)}~Initially, the quadrotor hovers on the right side of the pipe and a stable clockwise (CW) flow has developed. \textbf{b)}~As the quadrotor begins translating, the CW flow remains stable, but a small CCW vortex begins to form on the bottom right. \textbf{c)}~Due to the inertia of the flow, the quadrotor is clearly past the centerline before the circular flow breaks down. \textbf{d)}~A stable counter-clockwise (CCW) circular flow has developed. All photographs are taken with \unit[1/30]{s} exposure time.}
    \label{fig:flowstates_translation}
\end{figure*}

\section{Results}
\label{sec:results}

\subsection{Flow States}
\label{ssec:flow_states}
Before presenting quantitative results, we start by qualitatively explaining the flow states encountered in the pipe. 
Due to the long length of the pipe, there is no inflow or outflow, meaning that all air must be recirculated within the confinements of the pipe.

\subsubsection{Symmetric Configuration}
\label{sssec:symmetric_configuration}
\begin{figure}[t]
    \centering
    \includegraphics{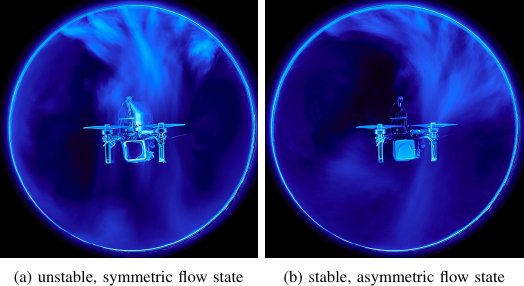}
    \vspace*{-12pt}
    \caption{Flow configurations when the quadrotor hovers in the center of the tube. \textbf{a)} The unstable symmetric flow is characterized by having two vortices of similar size to the left and to the right of the quadrotor. \textbf{b)} The flow configuration where one vortex is larger than the other one is observed to be stable. All photographs are taken with \unit[1/30]{s} exposure time.}
    \label{fig:flowstates_center}
\end{figure}

First, we consider the case where the quadrotor hovers in the center (e.g. ${y=0}$ and ${z=R/2}$), as shown in Fig.~\ref{fig:flowstates_center}. Since the setup is symmetric w.r.t. to~{$y=0$}, the flow can develop such that the downwash from the left propellers gets recirculated on the left in a clockwise vortex, and the downwash from the right propellers gets recirculated to the right in a counterclockwise vortex. This flow configuration is shown in Fig.~\ref{fig:flowstates_center}a.

However, this symmetric flow is not stable and can only be observed as a transient phenomenon. We observe that a flow configuration where one of the recirculation vortices is larger than the other one (see Fig.~\ref{fig:flowstates_center}b) is much more stable and\textemdash once developed\textemdash lasts for seconds. Looking at the image, this also intuitively makes sense: we see that the left, clockwise vortex is larger than its right counterpart. This deflects the downwash from both propellers to the left, thus sustaining the larger vortex.

\subsubsection{Asymmetric Configuration}
\label{sssec:asymmetric_configuration}
Next, we study the case where the quadrotor is hovering off-center, e.g. at~$y\neq 0$. Such a scenario is shown in Fig.~\ref{fig:flowstates_translation}a,d, where the quadrotor hovers to the right and left of the center, respectively. The flow that develops is a very stable, large-scale vortex encompassing the entire pipe. The measurements show that this circular flow can reach speeds up to \unit[5]{m/s}.

\subsubsection{Lateral Translation}
\label{sssec:lateral_translation}
Finally, we consider a dynamic scenario where the quadrotor starts out on one side of the pipe and laterally translates to the other side, as shown in Fig.~\ref{fig:flowstates_translation}. When the quadrotor is on the right initially (Fig.~\ref{fig:flowstates_translation}a), we observe the circular flow of the asymmetric configuration, and it remains stable as the quadrotor begins translating (Fig.~\ref{fig:flowstates_translation}b). 
When the quadrotor surpasses the centerline, an interesting phenomenon occurs: due to the inertia and the high velocity of the circular flow, it does not break down immediately. Only when the quadrotor is well past the centerline, as shown in Fig.~\ref{fig:flowstates_translation}c, the stable circular flow starts to break down. Briefly the asymmetric dual-vortex configuration seen previously when the quadrotor hovers in the center develops. Only \unit[1/4]{s} later, a stable, counter-clockwise circular flow has developed.
During this brief time interval, all disturbance torques and forces flip their sign, making the changeover from clockwise circular flow to counterclockwise circular flow the most challenging part of the flight.

\subsection{Optical Flow Estimation}
\label{ssec:optical_flow_estimation}
To evaluate the performance of our event-based smoke velocimetry (EBSV) method, we compare against both classical and learning-based optical flow estimation techniques. 
Since our method estimates sparse optical flow vectors at a fixed ${11 \times 11}$ grid of patches, we average the output of dense flow fields across the patches.

\subsubsection{Baselines and Groundtruth}
As a reference, we use flow fields computed with \emph{PIVLab}~\cite{thielicke2014pivlab, thielicke2021pivlab, thielicke2022motionblur}, a widely used open-source software for image-based particle image velocimetry. Specifically, we utilize the multiscale wavelet-based optical flow algorithm~\cite{wu1998waveletopticalflow, chen2002waveletopticalflow}, which provides dense velocity fields. We input the same event-frames used in our EBSV pipeline into PIVLab, as without a professional high-speed camera, no meaningful video could be obtained.

We also compare against a state-of-the-art unsupervised learning-based method (UL)~\cite{salah2022neuromorphicvisionbased} for optical flow estimation, originally designed for event data in real-world scenes, such as DSEC~\cite{gehrig21dsec} and MVSEC~\cite{zhu2018mvsec}. The method was retrained using our dataset to adapt to the smoke dynamics and motion statistics present in our domain.

As a second baseline, we implement a contrast maximization (CM)~\cite{gallago2018contrastmaximization} approach using variance as the contrast function. While contrast maximization has proven effective in scenes with sharp edges, its performance is expected to degrade in the presence of low-contrast, blob-like smoke structures.

\subsubsection{Quantitative Results}
Table~\ref{tab:optical_flow_methods} shows the results of our method and the selected baselines. We compare the RMSE (error here refers to the average endpoint distance of the flow vectors) on two sequences: the first is a slower sequence where the flow speeds are around \unit[2]{m/s} and the quadrotor is sitting on the ground while spinning its propellers, and the second is a faster in-flight sequence with speeds up to \unit[5]{m/s}. The real-time factor indicated in the table is $>1$ if the method runs faster than \unit[500]{Hz} real-time. For contrast maximization, we do not report a real-time factor, as our implementation was not optimized for runtime.

\begin{table}[ht]
    \centering
    \caption{Comparison of Optical Flow Methods on Smoke Data}
    \label{tab:optical_flow_methods}
\begin{tabularx}{1\linewidth}{l|CCC}
    \toprule
    Method & RMSE [m/s] \newline slow seq. & RMSE [m/s] \newline fast seq. & Real-Time Factor \\
    \midrule
    PIVLab & \textemdash & \textemdash & 0.0007 \\
    \midrule
    CM~\cite{gallago2018contrastmaximization} &  0.81 & 0.52 & \textemdash \\
    UL~\cite{salah2022neuromorphicvisionbased} & 1.18 & 0.7 & 0.27 \\
    Ours & \bf 0.34 & \bf 0.35 & \bf 3.2 \\
    \bottomrule
\end{tabularx}
\end{table}

As Tab.~\ref{tab:optical_flow_methods} shows, the proposed template matching method quantitatively outperforms the contrast-maximization and the unsupervised method while achieving the highest real-time factor. This clearly shows that well-known optical-flow estimators are not necessarily ideal for smoke velocimetry applications.

\subsection{Disturbance Wrench Estimation}
\label{ssec:disturbance_wrench_estimation}
After having demonstrated that our sparse optical flow estimation method is suitable for sparse velocimetry while being very low latency, the next step is to evaluate the disturbance estimation method presented in Sec.~\ref{sec:event_based_real_time_flow_estimation}. Due to the two-dimensional nature of the smoke velocimetry, we can only hope to recover disturbance information about the horizontal force $f^d_y$, the vertical force $f^d_z$, and the roll torque $\tau^d_x$. Consequently, we focus our evaluation in this section on those three quantities.

\subsubsection{Data Collection}
\label{sssec:data_collection}
To train and evaluate the disturbance wrench estimation network, a dataset comprising \unit[31]{min} of flight data is collected. To measure the disturbance wrench, we follow the methodology proposed in~\cite{bauersfeld2021neurobem}, e.g., combining onboard IMU and motor speed measurements with the state estimate from our monocular motion capture system. 
The dataset is split $2:1$ into training and test data, and all results presented in this section are based on the test split.

\subsubsection{A Position-Only Model}
To evaluate how important the real-time measurements of the flow are, we compare our full model with a simplified version consisting only of the position branch of our network architecture. 
This position-only model can capture all steady-state effects, and we expect it to perform well in cases where the time-varying components of the disturbance are small, e.g., when a stable CW or CCW circular flow has developed. 
In situations where the flow is unstable or has different possible configurations, the network will predict the average, as it is not able to capture multi-modality.

\begin{figure}[t]
    \centering
    \includegraphics{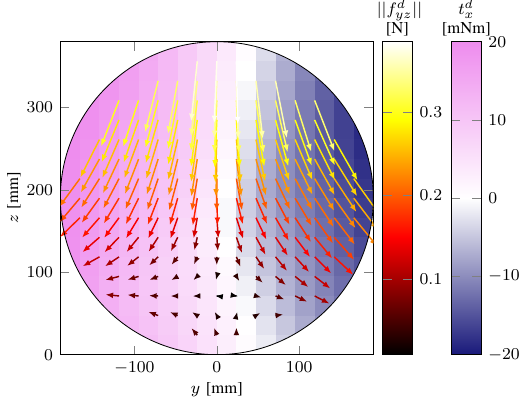}
    \vspace*{-12pt}
    \caption{The plot shows the predicted disturbances from the position-only model. The arrows indicate the forces; the colormap in the background represents the roll-torque disturbance. The measurements are symmetric w.r.t $y=0$, indicating no experimental bias.}
    \label{fig:position_wrench_network}
\end{figure}

\begin{figure*}[t!]
    \centering
    \includegraphics{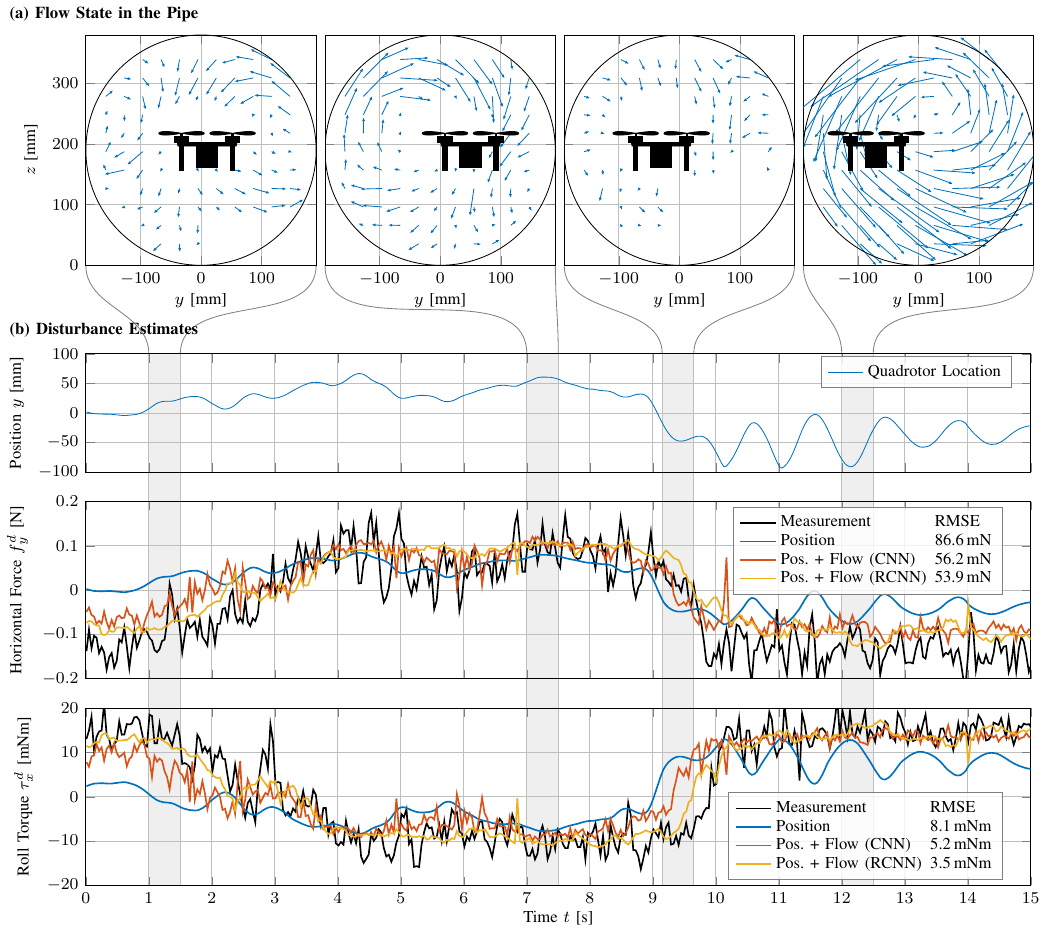}
    \caption{Smoke velocimetry and disturbance wrench estimation during a dynamic lateral translation. The quadrotor starts in the center, then first flies to the left half of the pipe, hovers there until the flow has developed, and then swiftly translates to the other side of the pipe and oscillates between the edge of the pipe and the center. \textbf{a)}~The sparse velocimetry estimates are similar to the qualitative images shown in Fig.~\ref{fig:flowstates_translation}. For improved clarity, we plot an average flow field over the \unit[0.5]{s} time windows indicated by the shaded areas. \textbf{b)}~The disturbance estimates for the horizontal force $f_y^d$ and roll torque $\tau_x^d$ clearly show that only the network using real-time flow estimates is able to capture the dynamic behavior during the quick translation across the pipe at $t=\unit[9.5]{s}$. The position-only baseline predicts the changeover to happen too early. We do not plot the disturbance in $f_z^d$ as both methods perform similarly.} 
    \label{fig:wrench_networks}
\end{figure*}

\begin{table}[p!]
    \vspace*{-6pt}
    \caption{Model Comparison for Disturbance Estimation (RMSE)}
    \label{tab:wrench_networks}
    \renewcommand{\arraystretch}{1.2}
\setlength{\tabcolsep}{4pt}
\begin{tabularx}{1\linewidth}{lX|>$c<$>$c<$>$c<$}
\toprule
Dataset & Model & f_y~[\unit{mN}] & f_z~[\unit{mN}] & \tau_x~[\unit{mNm}] \\ 
\midrule
\multirow{3}{*}{Overall}
 &     Position & 57.02 & 129.16 & 7.05 \\
 & Pos + Flow (CNN) & 44.46 & 126.94 & 5.27 \\
 & Pos. + Flow (RCNN) & \bf 41.32 & \bf 124.51 & \bf 4.96 \\
\midrule
\multirow{3}{*}{Steady-State}
 &     Position & 55.99 & 128.57 & 6.84 \\
 & Pos + Flow (CNN) & 44.10 & 126.55 & 5.31 \\
 & Pos. + Flow (RCNN) & \bf 40.77 & \bf 124.25 & \bf 4.97 \\
\midrule
\multirow{3}{*}{Dynamic}
 &     Position & 62.53 & 131.42 & 7.81 \\
 & Pos + Flow (CNN) & 47.27 & 128.44 & 5.13 \\
 & Pos. + Flow (RCNN) & \bf 45.45 & \bf 125.52 & \bf 4.92 \\
\bottomrule
\end{tabularx}
\end{table}

In Fig.~\ref{fig:position_wrench_network}, we plot the output of the position-only disturbance prediction. It can be seen that the aerodynamic disturbances both push the quadrotor towards the edge of the pipe and cause it to roll towards the outside of the pipe. 
This can be explained by considering the circular flow state: the flow velocity on the outside of the pipe is higher compared to the inside and consequently the pressure is lower on the outside, resulting in a force pointing outwards. 
Similarly, the higher flow speeds on the outside reduce the propeller efficiency of the outside propellers, resulting in a roll-torque disturbance. 
In the $z$-direction, we observe a negative disturbance force, indicating reduced lift due to the non-zero inflow velocity above the rotor disk. 
Only close to the bottom of the pipe does the ground effect compensate for this loss of lift. 
Lastly, we notice that the plot is symmetric w.r.t. $y=0$, indicating that the de-biasing during training is effective, and no general bias in the experimental data remains.

\subsubsection{Position \& Flow Model}
\label{sssec:distwrench_pos_flow_model}
Our full model presented in Sec.~\ref{ssec:wrench_estimation} utilizes both the ConvLSTM branch to process the flow measurements, as well as the position branch to account for the position of the quadrotor. 
This model is referred to as \emph{Pos. + Flow (RCNN)}.
In addition to the recurrent ConvLSTM version, we also evaluate a non-recurrent variant where the ConvLSTM layers are replaced with standard 2D convolutions, labeled as \emph{Pos. + Flow (CNN)}.
We evaluate the networks in terms of disturbance-wrench prediction RMSE on the test split, and the results are presented in Tab.~\ref{tab:wrench_networks}.

Considering all test data, the estimation of the disturbance $f^d_y$ and $\tau^d_x$ degrades by \unit[38]{\%} and \unit[42]{\%}, respectively, when not using the flow estimates for disturbance estimation, highlighting the usefulness of real-time smoke velocimetry in this scenario. 
For $f_z^d$, the improvement is only \unit[4]{\%}, indicating that potentially relevant flow structures might not be captured by our sparse velocimetry approach. 
However, considering that the $z$-direction on a quadrotor is directly actuated, and that a prediction error of \unit[0.12]{N} RMS on a quadrotor requiring \unit[2.14]{N} static hover thrust is only \unit[6]{\%}, this prediction accuracy is likely sufficient for control. 
We also observe that adding recurrency to the flow processing part of the network helps, as the RCNN is between \unit[2]{\%} and \unit[6]{\%} better in disturbance estimation compared to the CNN variant.

Next, we split the test dataset into steady-state sequences and dynamic sequences, depending on whether the quadrotor moved more than \unit[5]{cm} laterally. From Tab.~\ref{tab:wrench_networks}, we can see that the overall accuracy degrades slightly, but the difference between the position-only and the full input becomes more pronounced in dynamic sequences, with up to \unit[59]{\%} difference in the roll disturbance. This highlights the importance of flow sensing in dynamic scenarios.

\begin{figure}[t!]
    \centering
    \includegraphics{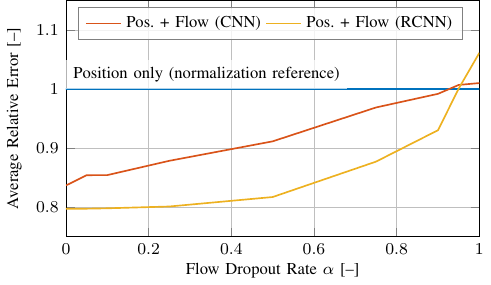}
    \vspace*{-6pt}
    \caption{Comparison of the average normalized errors of the disturbance estimator under increasing dropouts of flow measurements. The error is normalized w.r.t. the error of the \emph{Position} model and averaged across $f_y$, $f_z$, and $\tau_x$. Our RCNN architecture maintains its performance up to \unit[25]{\%} dropouts of flow measurements.}
    \label{fig:wrench_networks_dropout}
    \vspace*{-6pt}
\end{figure}

\rebuttal{
Finally, we analyze the performance of the disturbance estimator under flow measurement dropouts with dropout probability $\alpha$ focusing on the case of completely dropped (e.g. zeroed) measurements across entire frames. Figure~\ref{fig:wrench_networks_dropout} shows the performance of the networks using flow information relative to the position-only network. We see that, in the absence of flow measurements (i.e. $\alpha=1$), the performance of the networks is marginally worse compared to position-only. Our recursive RCNN architecture shows good robustness against dropouts, with the accuracy being practically unaffected until \unit[25]{\%} random dropouts, whereas the non-recurrent CNN immediately shows a decrease in performance with dropped measurements.
}

\subsubsection{Lateral Translation}
Finally, we revisit the lateral translation maneuver introduced in Fig.~\ref{fig:flowstates_translation} and analyze the performance of the disturbance estimation models in terms of lateral force $f_y^d$ and roll torque $\tau_x^d$. Figure~\ref{fig:wrench_networks} provides a detailed comparison between the ground truth disturbances and the estimates produced by the position-only model and the CNN\,/\,RCNN models augmented with flow input.

At $t = \unit[0]{s}$, the quadrotor is hovering near the centerline of the pipe. However, despite this apparently symmetric location, the underlying flow field is not perfectly symmetric. Specifically, a stronger counter-clockwise (CCW) vortex has developed to the right of the vehicle, while the clockwise (CW) vortex on the left is smaller. This asymmetric flow field induces a non-zero lateral force and roll torque, even though the quadrotor is centered. The position-only model, lacking access to the actual flow conditions, fails to predict this accurately. In contrast, the models with real-time flow input successfully capture the asymmetry and produce more accurate disturbance estimates.

As the quadrotor translates towards one side of the pipe and remains there, the flow settles into a fully developed circular pattern ${t = \unit[7]{s}}$. Under these steady-state conditions, all models perform well in estimating the resulting disturbance wrench. 

The key differences between the models emerge again during dynamic transitions. Around $t = \unit[9.5]{s}$, the quadrotor performs a lateral translation across the pipe. 
During this motion, the flow structure undergoes a reconfiguration, transitioning from CW to CCW circular flow. 
The position-only model incorrectly predicts the sign change in disturbances (particularly torque) to occur exactly as the quadrotor crosses the pipe’s centerline. 
In contrast, the flow-augmented models are more accurate, and especially the RCNN version delays the predicted sign reversal until after the quadrotor has passed the center. 
This hysteresis-like behavior is consistent with the analysis shown earlier in Fig.~\ref{fig:flowstates_translation}c and highlights the limitations of relying solely on geometric information.

Finally, once the CCW flow is fully established again (after $t = \unit[10]{s}$), the flow field stabilizes. 
In this regime, even positional oscillations of the quadrotor do not significantly alter the actual aerodynamic wrench. 
The models with flow input reflect this stability by producing disturbance predictions that remain consistent despite the oscillations. 
Conversely, the position-only model is inherently unable to predict this, and erroneously associates every change in position as a change in disturbance force or torque. 

These findings underscore the value of incorporating real-time flow measurements into the disturbance estimation process. While static relationships between position and flow-induced disturbances can sometimes suffice, accurate prediction during transient and asymmetric flow conditions requires real-time observation of the flow state. Additionally, we find that incorporating recurrency into the wrench estimation leads to more accurate and less noisy estimates.

\subsection{Closed-Loop Flight Performance}
\label{ssec:flight_performance}
A central objective of this work is to demonstrate how real-time event-based image velocimetry can be leveraged within a closed-loop control system. 

\subsubsection{Hovering Flight}
\label{sssec:hovering_flight}
To this end, we first analyze the performance of different disturbance-handling strategies during hovering flight, as summarized in Table~\ref{tab:closed_loop_hover}.
We evaluate three policies trained under different configurations. The first, referred to as \emph{baseline}, is trained without exposure to any aerodynamic disturbances and with zeroed-out disturbance observations. The two other policies, \emph{wo/ dist. obs.} and \emph{w/ dist. obs.}, are trained with the full aerodynamic disturbance simulation as described in Sec.~\ref{ssec:training_in_simulation}.
The difference between the two is that the \emph{wo/ dist. obs.} policy has zeroed-out disturbance observations during training. The policy \emph{w/ dist. obs. (ours)} also has access to real-time information about the disturbance during training. 

In Tab.~\ref{tab:closed_loop_hover}, the standard deviation and inter-quartile ranges of the hover position are shown. We observe that incorporating disturbance modeling during training improves hover accuracy, as both the \emph{wo/ dist. obs.} and \emph{w/ dist. obs.} policies perform much better than the baseline.
This improvement is likely due to the consistent aerodynamic wrenches present in the confined environment, which tend to push the quadrotor outward from the centerline. 
A controller trained under such conditions learns to anticipate and compensate for this instability. Furthermore, comparing the second and third policies, we find that access to real-time disturbance observations further enhances control performance. 
Specifically, our policy improves position standard deviation in the $y$-direction by \unit[29]{\%}, demonstrating the value of integrating real-time event-based flow information into the control pipeline. 
\rebuttal{
As described in Sec.~\ref{sssec:lstm_ppo_policy_architecture} we simulate measurement dropouts during training. Our experiments empirically validate this approach: if no smoke is present in the pipe the \emph{w/ dist. obs. (ours)} policy has no real-time disturbance observations and it performs on par with its \emph{wo/ dist. obs.} counterpart.
}

\begin{table}[t!]
    \centering
    \vspace*{-12pt}
    \caption{Hover Position Deviation (Flight in Center)}
    \label{tab:closed_loop_hover}
     \begin{tabularx}{\linewidth}{p{2.5cm}|CCCC}
    \toprule
    Model & \multicolumn{2}{c}{Pos. Y [mm]} & \multicolumn{2}{c}{Pos. Z [mm]} \\
    & std. dev. & IQR & std. dev. & IQR \\ \midrule
    baseline & 34.2 & 44.3 & 7.7 & 10.9 \\ 
    wo/ dist. obs. & 14.3 & 17.3 & 12.4 & 17.5 \\ 
    w/ dist. obs. (ours) &  \bf 10.3 & \bf 14.9 & \bf 11.3 & \bf 17.4 \\
    \bottomrule
\end{tabularx}
    \vspace*{-12pt}
\end{table}

\subsubsection{Lateral Translation}
\begin{figure}
    \centering
    \includegraphics{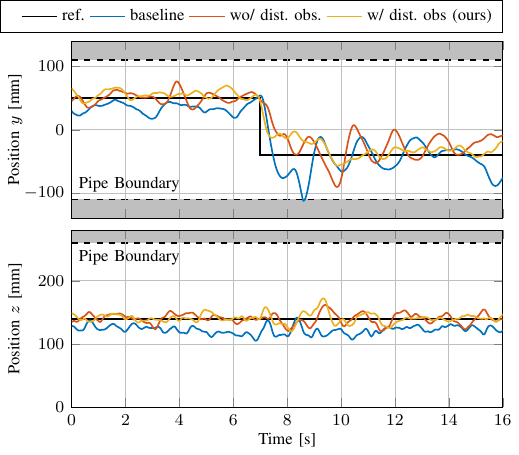}
    \vspace*{-20pt}
    \caption{Position in $y$ and $z$ of the quadrotor during lateral translation across the pipe. The pipe boundary shows when the quadrotor would hit the wall. We compare a policy trained without any aerodynamic disturbances \emph{baseline}, a policy that is subjected to disturbances during training but has no observation of them \emph{wo/ dist. obs.}, and a policy using our real-time disturbance feedback \emph{w/ dist. obs. (ours)}. The policy trained without disturbances crashes into the pipe wall. Using real-time disturbance estimation reduces the overshoot by \unit[71]{\%} compared to the baseline that has no access to real-time measurements.}
    \label{fig:closed_loop_translation}
    \vspace*{-12pt}
\end{figure}

While improvements during hover are observable, the advantages of flow-aware disturbance estimation become even more apparent in dynamic flight scenarios. 
To investigate this in detail, we analyze a lateral translation maneuver in which the quadrotor is commanded to move from \unit[50]{mm} to \unit[-40]{mm} in the $y$-direction. 
The resulting position trajectories in $y$ and $z$ of one experiment are shown exemplarily in Fig.~\ref{fig:closed_loop_translation}. 
During this maneuver, the quadrotor must remain within \unit[$\pm 110$]{mm} in the $y$-direction to avoid crashing into the pipe wall.

We again compare the three previously described policies and repeat the experiment 5 times.
The baseline policy fails to complete the translation safely, colliding with the wall shortly after the flow direction reverses in all cases. 
It does not stabilize thereafter and exhibits sustained oscillations. The policy trained without disturbance observation performs better; it avoids wall contact and stabilizes more quickly at the target position, although some overshoot remains. 
In contrast, our policy with disturbance observation successfully completes the maneuver without overshooting, maintaining a safe margin throughout. On average, we find that including flow measurements in the observation leads to \unit[71]{\%} less overshoot than \emph{wo/ dist. obs.}.

In the $z$-direction, the performance difference among the policies is less pronounced. 
Both disturbance-trained policies are generally capable of maintaining the altitude setpoint of \unit[140]{mm}. 
Only the policy trained without any aerodynamic forces hovers too low because it is not used to the reduced lift due to recirculation.

\section{Discussion \& Limitations}
\label{sec:discussion}

The results presented in the previous section successfully demonstrate the use of real-time, event-based smoke velocimetry for disturbance-aware closed-loop control of a quadrotor operating in confined environments. The control performance is substantially improved through the integration of low-latency flow feedback into the control pipeline. However, this is not the first time flight in narrow pipes has been demonstrated, and this section puts our results into context and discusses remaining limitations.

\vspace*{-6pt}
\subsection{Context with Related Works}
\label{ssec:context_w_rel_work}
The quadrotor and the pipe used in~\cite{martin2025flyinginducts} are similar to our flight hardware setup and thus the results can be meaningfully compared, with the main difference being the difference in quadrotor mass (\unit[130]{g} theirs, \unit[210]{g} ours).
In contrast to our indirect force/torque sensing, they use a robotic arm with a 6-DOF wrench sensor to place the quadrotor at various positions in the pipe and record the aerodynamic forces. 
While it is reassuring that their method yields qualitatively similar results, their approach suffers from two problems. 
First, the quadrotor does not have to counteract the aerodynamic disturbances it creates, as it is held by the robot arm. 
Second, the use of the arm requires them to position the quadrotor close to the inlet of the pipe, making it impossible to consider the results accurate for an infinite pipe. 
We believe that this is a reason why their force estimates are \unit[30]{\%} to \unit[50]{\%} less than ours (even after scaling everything w.r.t. the quadrotor mass), because the flow can leave the pipe.

Looking at the control strategy, they have no real-time feedback on the flow and thus, method-wise, our control policy trained without disturbance observation is comparable. 
For hovering in the center of the pipe, they report an inter-quartile range (IQR) of about \unit[70]{mm} in the $y$ direction (read off from~\cite{martin2025flyinginducts} Fig.~5b). 
Even without flow feedback, we obtain a much lower IQR of around \unit[17.3]{mm}, which is reduced even further to \unit[14.9]{mm} with the flow feedback. 
This demonstrates the superior performance of our LSTM-PPO controller over the cascaded PID architecture in~\cite{martin2025flyinginducts}.

When we compare our results with~\cite{wang2025autonomousflighttunnel}, we must first note that this is not a one-to-one comparison because of the differently sized drones.
Nothing about the scaling laws of quadrotor-induced recirculation flows in pipes is known and, consequently, the results cannot be directly compared. 
Their work is a milestone in itself, as they do not rely on external state estimation and perform all required computations onboard, enabling true autonomy. 
However, this autonomy comes at the expense of being unable to hover\textemdash a maneuver critical in many inspection scenarios. 
Our insights into the flow configurations and the temporal structure of the disturbances could potentially be integrated into a fully autonomous system, improving its performance.

\vspace*{-7pt}
\subsection{Limitations}
\label{ssec:limitations}
When we look at this work from an industrial application context, the main limitation is the need for an instrumented environment. 
Our approach is tailored specifically for operation inside a narrow pipe equipped with a smoke injection system and an external light sheet for flow visualization. Additionally, state estimation is done with our external monocular event-based motion capture.
This constrains the applicability of the method to environments where such experimental infrastructure is feasible and rules out deployment in more general or unstructured settings. 
However, from a scientific point of view, this is not a limitation but a key enabler to better understand the recirculation flow and aerodynamic disturbances that\textemdash without the smoke and lightsheet\textemdash would be unobservable.

A second limitation comes from the hardware itself. Although the overall, worst-case latency of the flow estimation system is extremely low and on the order of \unit[1-2]{ms}, the full control loop includes additional delays due to wireless communication with the quadrotor and actuator dynamics.
These introduce a bottleneck that prevents the system from fully exploiting the speed of the event-based flow measurements. 
As a result, there is a delay between disturbance observation and quadrotor response that ultimately limits the reactivity of the controller and thus the maximal achievable disturbance compensation.

\vspace*{-8pt}
\subsection{Event-Camera vs. High-Speed Camera}
\label{ssec:eventcamera_vs_highspeed}
Lastly, we want to discuss the choice of an event camera instead of a more standard high-speed camera for flow sensing.
At the same resolution and a flow speeds of \unit[5]{m/s}, we would need a 1/2500–1/4000s exposure time to limit motion blur to below \unit[2]{px}. 
To assess the feasibility, we first calculate the luminance of the scene based on the photos~\cite{ISO2720-1974}:
\begin{equation}
L = (A^2 / T) \cdot (K / S) = \unit[3.5]{cd/m^2}
\end{equation}
where $L$ is the luminance, $A$ the aperture (f-stop number), $T$ the exposure time, $S$ the ISO sensitivity, and $K$ a calibration factor in the range of 10.6 to 13.4~\cite{ISO2720-1974}. We use $K=\unit[12.5]{cd\,s/m^2}$ and obtain a fairly low luminance on the order of $10^0$. This luminance level already requires twilight-vision in humans~\cite{cie1978luminance}. To achieve the shutter speeds around 1/2500-1/4000 in a conventional high-speed camera, this would amount to ISO 25600 with an f/1.4 lens. Only the most advanced high-speed cameras can even achieve such high sensitivity, but the image quality will be severely degraded due to noise.

Therefore, achieving sufficient illumination for high-speed standard imaging inside the pipe would ideally require a pulsed laser sheet, which is impractical due to safety and engineering constraints. 
Furthermore, most high-speed cameras are not optimized for machine vision and do not provide a low-latency image stream as output.
Event cameras, on the other hand, provide high temporal resolution and low latency under the same conditions, at significantly lower cost and complexity. 

The sparse event-stream also provides significant advantages in terms of computational load compared to frame-based high-speed video cameras. This enables running our event-based motion-capture on small embedded computers such as the Raspberry Pi 4B. Similarly, the event-based smoke velocimetry can be deployed on Nvidia's Jetson family, enabling future applications where onboard deployments are required. 
Designing a larger drone platform which tightly integrates the light sheet and smoke generation system, a lightweight event camera and a Jetson computer into a single platform would be a next step allowing real-time flow sensing to be performed in real-world environments without relying on an instrumented pipe.
\vspace*{6pt}

\section{Conclusion}
\label{sec:conclusion}
In this work, we present the first demonstration of real-time, event-based flow field measurements used for closed-loop control of a quadrotor in a narrow pipe. We develop a sparse, low-latency, event-based smoke velocimetry pipeline and use it to train a recurrent convolutional disturbance estimator that outputs aerodynamic wrench estimates in real time. These estimates are then integrated into a reinforcement learning-based controller trained to compensate for disturbances during flight.

Our results show that this approach significantly improves performance in both hovering and dynamic flight scenarios, reducing positional error by up to \unit[29]{\%} and overshoot by \unit[71]{\%} compared to models that lack flow-based feedback. This work demonstrates not only the feasibility of closed-loop, flow-aware control for aerial robots but also sets a precedent for integrating event-based perception and fluid sensing into robotic systems operating in aerodynamically complex environments.

To the best of our knowledge, this is the first time that closed-loop control with real-time flow field feedback has been demonstrated on an aerial robot. We believe this opens new directions for research on robust flight in turbulent and confined environments where standard sensing and control approaches are insufficient.  Additionally, understanding how characteristic flow patterns and disturbance magnitudes depend on the pipe’s cross-section and the drone’s geometry and mass offers an exciting avenue for future fluid dynamics research.

\clearpage
{
\bibliographystyle{IEEEtran}
\bibliography{references}
}

\appendices
\section{Blinking LED Circuit}
\label{sec:blinking_led_circuit}
\subsection{Choice of LEDs}
\label{subsec:choice_of_leds}
For the accuracy of the event-based mocap system, it is important that the center of the blinking LED can be detected easily by the event camera. To achieve this, the LED should be small, very bright, and have a short switching time to produce a well-defined rising and falling edge. These requirements are ideally met by LEDs optimized for pulsed operation, such as infrared LEDs for data transmission.

In this work, the Osram SFH4350~\cite{osram4350} IR-LED \unit[3]{mm} was selected, as it features an extremely short switching time of \unit[12]{ns}. Being designed for optoelectronics, it also permits up to \unit[1]{A} pulsed forward current for pulses shorter than \unit[100]{$\mu$s}, if the duty cycle is below \unit[2]{\%}. The LED has its peak emission around \unit[850]{nm}, with a spread of \unit[50]{nm}, which is within the sensitivity range of most CMOS-based imaging sensors.

\vspace*{-9pt}
\subsection{Circuit Design}
\label{subsec:circuit_design}
The design of the LED driver circuit is tightly coupled with the entire event-based motion capture system:
\begin{enumerate}
    \item Faster blinking frequencies increase the responsiveness of the system, as at least one full period must be detected to identify an LED. For robustness, detecting at least two periods is preferable. Consequently, if the LEDs blink at \unit[1]{kHz}, the overall system is limited to a \unit[500]{Hz} output rate.
    \label{item:lower_limit}
    \item If the LEDs blink too fast, measuring the signal becomes difficult. Typically, event cameras perform very well at measuring signals with frequencies up to \unit[2-3]{kHz}~\cite{wang2024towardshighspeed, crabtree2023refactoryperiod, lichtsteiner2008a128x128dvs}, and the events are timestamped with \unit[1]{$\mu$s} time resolution.
    \label{item:upper_limit}
    \item At least four LEDs are necessary to yield a unique solution to the PnP problem.
    \label{item:number_leds}
    \item The individual frequencies should not alias into each other. This means that, ideally, all LEDs have blinking frequencies within a factor of two.
    \label{item:aliasing}
    \item Due to the limitations of the LED, a duty cycle of \unit[2]{\%} cannot be exceeded.
    \label{item:duty}
\end{enumerate}

To control the blinking LEDs, either a microcontroller or an analog circuit can be used. Because the high LED forward current of \unit[1]{A} necessitates an analog output stage, we opted for a fully analog design using NE555~\cite{ne555} precision timers. To generate the signal for the LEDs, the NE555 is operated in astable mode (cf. Sec. 8.3.2~\cite{ne555}). 

The current output of the precision timer is limited to \unit[200]{mA}, but its performance significantly degrades if the output current exceeds \unit[10]{\%} of the maximum value (cf. Figure 3 of~\cite{ne555}). Therefore, an SPST (single pole, single throw) digital switch is used. We selected the ADG802, as it supports close to \unit[1]{A} pulsed current and has typical switching times around \unit[55]{ns}~\cite{adg802}. While considerably slower than the LED, it is still fast enough for the given application.

\begin{table}[t!]
    \centering
    \caption{\vspace*{6pt}\parbox{1\linewidth}{\textnormal{Resistor and capacitor values for the different LEDs. The calculated periods as well as the measured frequencies $f$ and duty cycles $\alpha$ are listed.}}}
    \vspace*{-6pt}
    \label{tab:resistor_capacitor_values}
    \begin{tabularx}{1\linewidth}{CCC|CCC|CC}
\toprule
\multicolumn{3}{c|}{Part Specification} & \multicolumn{3}{c|}{Calc. from Sec. 8.3.2 \cite{ne555} } & \multicolumn{2}{c}{Measured} \\[2pt]
$R_A$ \newline [$\unit{k\Omega}$] 
& $R_B$ \newline [$\unit{k\Omega}$] 
& $C$ \newline [\unit{nF}] 
& $t_\text{on}$ \newline [$\unit{\mu s}$] 
& $t_\text{off}$ \newline [$\unit{\mu s}$] 
& $f$ \newline [\unit{kHz}]
& $f_\text{meas}$ \newline [\unit{kHz}] 
& $\alpha_\text{meas}$ \newline [\%]\\ \midrule %
68.1 & 0.39 & 10 & 2.7 & 477 & 2.094 & 1.73 & 0.66 \\
59.0 & 0.39 & 10 & 2.7 & 415 & 2.413 & 1.98 & 0.75 \\
51.1 & 0.39 & 10 & 2.7 & 359 & 2.781 & 2.29 & 0.87 \\
44.2 & 0.39 & 10 & 2.7 & 312 & 3.207 & 2.61 & 0.99 \\
40.2 & 0.39 & 10 & 2.7 & 284 & 3.520 & 2.86 & 1.09 \\
\bottomrule
\end{tabularx}
\end{table}

\begin{figure}[t!]
    \centering
    \includegraphics{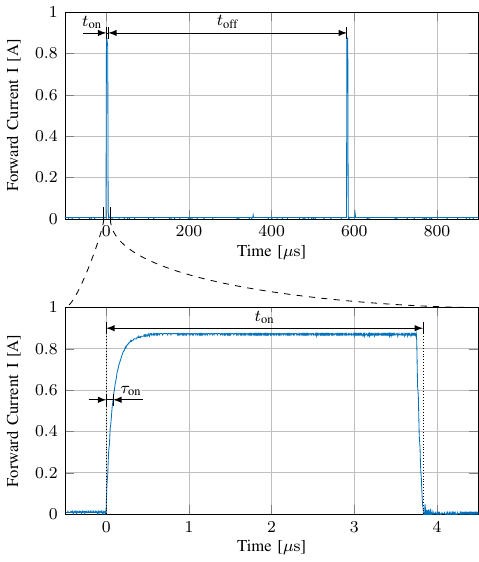}
    \vspace*{-9pt}
    \caption{Current through an LED blinking at the lowest frequency of \unit[1.73]{kHz}. The upper plot shows the entire period $t = t_\text{on} + t_\text{off}$, whereas the lower plot shows only the pulse during which the LED is on for \unit[3.8]{$\mu$s}. From the plot, we get a switch-on time constant $\tau_\text{on}$ of the LED of \unit[84]{ns} (time to reach \unit[63]{\%} of the steady-state value).}
    \label{fig:led_current}
    \vspace*{-12pt}
\end{figure}

The LEDs, precision timers, and switches are all supplied with a single LM7805 voltage regulator. This is possible because the time-averaged load is well below the design limit of the voltage regulator. Each NE555 draws {$I_\text{NE555} = \unit[3]{mA}$} of supply current. The time-averaged current $\bar{I}_\text{LED}$ for an LED pulsed with a duty cycle $\alpha = t_\text{on} / t_\text{period}$ with a pulse current {$I_p = \unit[1]{A}$} is given by
\begin{equation}
    \bar{I}_\text{LED} = \alpha \cdot I_p
\end{equation}
Assuming there are $N=5$ LEDs and they are operated at an average duty cycle of \unit[1]{\%}, this leads to a total time-averaged current of 
\begin{equation}
    \bar{I} = N \cdot \left(\bar{I}_\text{LED} + I_\text{NE555} \right) = \unit[65]{mA}
\end{equation}
which is within specifications for an LM7805~\cite{lm7805} without any additional cooling (given the TO220 package and a \unit[12]{V} supply). Therefore, a large decoupling capacitor is used to supply the LEDs with power, effectively shielding the LM7805 from all current spikes caused by the LEDs. Based on all the above considerations, the circuit has been designed and subsequently manufactured into a PCB with SMD versions of the NE555 and the ADG802.

The resistor and capacitor values used in the NE555 timer circuit are listed in Tab.~\ref{tab:resistor_capacitor_values}. The values have been calculated such that the frequencies of all 5 LEDs satisfy points \ref{item:upper_limit} to \ref{item:duty}. After building and manufacturing the PCB, the measured frequencies are also listed in Tab.~\ref{tab:resistor_capacitor_values}. Note that the mismatch with respect to the calculated values is about \unit[20]{\%}. This mismatch is consistent across 5 identical copies of the board and is of no concern for practical applications, as it is straightforward to measure the blinking frequency with an oscilloscope. Exemplarily, Fig.~\ref{fig:led_current} shows the current through one LED.

\section{Pose Estimation Noise}
\label{sec:pose_estimation_noise}
\begin{figure}
    \centering
    \includegraphics{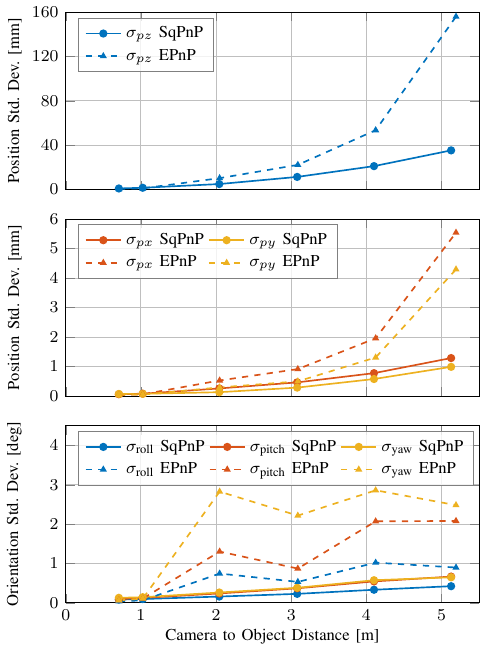}
    \caption{The object is placed at distances between \unit[70]{cm} and \unit[5]{m} statically in front of the camera (with the \unit[25]{mm} lens). The plots show the standard deviation in the position measurement ($z_\cfr$ and $x_\cfr$, $y_\cfr$) as well as the orientation measurements. We can clearly see that SqPnP~\cite{terzakis2020sqpnp} outperforms EPnP~\cite{lepetit2009epnp} by a large margin.}
    \label{fig:pose_noise}
    \vspace*{-7pt}
\end{figure}

To measure the pose estimation noise, the drone is rigidly placed at various distances in front of the camera. Then, 10 seconds of data are recorded with the event camera and we calculate the standard deviation of the pose estimate. Since the drone is static, all deviations from the mean are only due to noise. The results of this analysis are summarized in Fig.~\ref{fig:pose_noise}. A few interesting observations can be made, which are subsequently discussed in detail:
\begin{enumerate}
    \item The noise levels along the $z_\cfr$ axis are much larger compared to the $x_\cfr$ and $y_\cfr$ axes.
    \item The SqPnP~\cite{terzakis2020sqpnp} algorithm achieves much better performance than EPnP~\cite{lepetit2009epnp}.
    \item For SqPnP, the noise levels in position scale quadratically with the distance from the camera, while the orientation noise scales linearly.
\end{enumerate}

The $z_\cfr$-axis is the optical axis of the camera, and hence the $z_\cfr$ coordinate can only be inferred from the scale of the object. Assuming that the elongation of the object along the optical axis is small compared to the distance to the camera (i.e., the object is nearly flat), a well-known result from stereo vision applies~\cite{zisserman2004multipleview}: for a given inter-marker distance $d$, focal length $f$, and a marker detection with uncertainty $\sigma_u$ (in pixels), the depth uncertainty $\sigma_{pz}$ scales with the square of the distance $z$ as
\begin{equation}
    \sigma_{pz} = \frac{\partial z_\cfr}{\partial u} \cdot \sigma_u = \frac{b \cdot f}{z_\cfr^2} \cdot \sigma_u \sim \frac{1}{z^2}\;.
\end{equation}

For the positional errors in $x_\cfr$ and $y_\cfr$, we observe a similar quadratic dependency on the camera-object distance, however with much less noise. Intuitively, this makes sense, as the translation along $x_\cfr$ and $y_\cfr$ is directly observable from each marker and thus the estimate is much more accurate.

The position noise plots also highlight the superior performance of SqPnP for this task: the optimization-based approach is able to estimate the position with much less variance given the same input data. This discrepancy becomes even larger when considering the orientation estimation shown in the bottom plot of Fig.~\ref{fig:pose_noise}. SqPnP dramatically outperforms EPnP, which performs between two and four times worse. Interestingly, we observe that EPnP shows a large but nearly constant orientation uncertainty after \unit[2]{m}, whereas SqPnP shows a linear increase in the noise standard deviation. Note that we do not compare against EPnP with nonlinear refinement, as the OpenCV implementation requires at least six points for iterative refinement.

\begin{figure}
    \centering
    \includegraphics{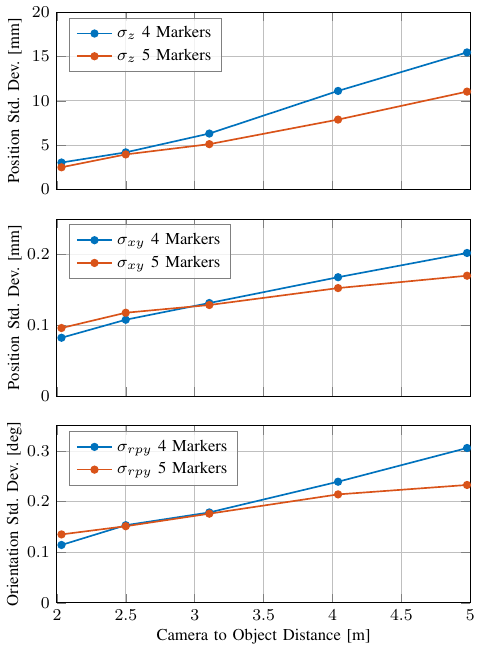}
    \caption{
    Effects of marker occlusion on the SqPnP accuracy. The drone is placed at different positions along the pipe (pipe center is at a distance of \unit[3]{m}). Comparing the pose estimation error between the occluded and the unoccluded case we see that the effects of occlusion are very well handled by SqPnP.
    }
    \label{fig:marker_occlusion}
    \vspace*{-7pt}
\end{figure}

In addition to the above comparison between EPnP and SqPnP, we also analyze the performance of SqPnP when a single marker is occluded and thus only $N=4$ markers are visible. The experiments are conducted inside the pipe with the Zeiss \unit[50]{mm} lens under identical conditions to those used for flight experiments. Fig.~\ref{fig:marker_occlusion} shows the longitudinal ($z_\cfr$) and lateral position error as well as the average attitude error. Overall, the difference between the 4 marker and 5 marker case is small and, from a practical perspective, not relevant. Interestingly, in the longitudinal direction we observe that using 4 vs. 5 markers strictly increases the measurement noise, whereas at close distances the measurement noise parallel to the image plane and in the attitude is slightly lower for 4 markers. This is counterintuitive at first, but could be explained if we consider imperfect 3D positions of the LEDs on the drone. With 5 markers the solver has more ability to 'jump' between solutions, increasing the noise level.

\end{document}